\documentclass{article}

\usepackage[utf8]{inputenc} %
\usepackage[T1]{fontenc}    %
\usepackage{hyperref}       %
\usepackage{url}            %
\usepackage{booktabs}       %
\usepackage{amsfonts}       %
\usepackage{nicefrac}       %
\usepackage{microtype}      %
\usepackage{xcolor}         %

\usepackage{amsmath}
\usepackage{xcolor}
\usepackage[normalem]{ulem}
\usepackage{algpseudocode}
\usepackage{amsfonts}
\usepackage{graphicx}
\usepackage{adjustbox}
\usepackage{amssymb}
\usepackage{algorithm}
\usepackage{pdfpages}
\usepackage{multicol}
\usepackage{multirow}
\usepackage{wrapfig}
\usepackage{lipsum}

\newtheorem{definition}{Definition}

\newtheorem{res-quest}{Research Question}

\definecolor{ao(english)}{rgb}{0.0, 0.5, 0.0}

\definecolor{greentxt}{HTML}{00A64F} 
\definecolor{realpurple}{HTML}{BD33A4} 

\newcommand*{\icon}[1]{%
    \raisebox{-.1\baselineskip}{%
        \includegraphics[
        height=0.6\baselineskip,
        width=0.6\baselineskip,
        keepaspectratio,
        ]{#1}%
    }%
}

\title{\textbf{Emergent Linguistic Structures in Neural Networks are Fragile}}

\author{Emanuele La Malfa\thanks{Department of Computer Science,
        University of Oxford, Oxford, United Kingdom. Corresponding address: emanuele.lamalfa@cs.ox.ac.uk}        
    \and    
    Matthew Wicker\textsuperscript{*}\thanks{This work has been done while the author was affiliated with the University of Oxford. Current affiliation: The Alan Turing Institute, London, United Kingdom}
    \and
    Marta Kwiatkowska\textsuperscript{*} 
    }%

\begin{document}

\maketitle
\begin{abstract}
\noindent Large Language Models (LLMs) have been reported to have strong performance on natural language processing tasks. However, performance metrics such as accuracy do not measure the quality of the model in terms of its ability to robustly represent complex linguistic structures. In this paper, focusing on the ability of language models to represent syntax, 
we propose a framework to assess the consistency and robustness of linguistic representations. To this end, we introduce measures of robustness of neural network models that leverage recent advances in extracting linguistic constructs from LLMs via probing tasks, i.e., simple tasks used to extract meaningful information about a single facet of a language model, such as syntax reconstruction and root identification. 
Empirically, we study the performance of four LLMs across six different corpora on the proposed robustness measures by analysing their performance and robustness with respect to syntax-preserving perturbations. We provide evidence that context-free representation (e.g., GloVe) are in some cases competitive with context-dependent representations from modern LLMs (e.g., BERT), yet equally brittle to syntax-preserving perturbations. Our key observation is that emergent syntactic representations in neural networks are brittle. We make the code, trained models and logs available to the community as a contribution to the debate about the capabilities of LLMs.  

\end{abstract}

\section{Introduction} \label{sec:intro}

Large Language Models (LLMs) %
exhibit impressive performance in natural language processing (NLP) tasks such as text classification~\cite{liu2019roberta}, language translation~\cite{liu2020very}, %
large-scale search engines~\cite{wang2021dcn}, and even source-code generation for programming languages \cite{chen2021evaluating}, resulting in a great deal of media attention. %
However, current metrics mainly %
measure the performance of LLMs' ability to capture statistical patterns of discourse~\cite{bender2021dangers,manning1999foundations}, as opposed to their ability to robustly capture and represent complex linguistic patterns of their domains. 

Representing the linguistic and grammatical structure underlying the data intuitively plays a cogent role in robust generalization of any linguistic system~\cite{chomsky2006language}. Remarkably, LLMs are also arguably capable of accurately representing structures such as syntax trees~\cite{manning2020emergent}, which has motivated researchers to investigate their linguistic capabilities with ad hoc measures and benchmarks~\cite{rogers2020primer,sinha2021masked}. 
This %
indisputable progress is nonetheless counteracted by a series of critiques that show how LLMs are unable to perform basic reasoning \cite{niven2019probing}, have considerable biases \cite{le2020adversarial}, are not well aligned to stakeholder values \cite{bender2021dangers}, and are brittle in the face of adversarial examples \cite{la2021king}. Such studies make it clear that sustainable, long-term advances in NLP need to be facilitated by appropriate %
metrics that capture how LLMs represent the complex linguistic patterns underlying their training data \cite{bender2020climbing}. Unfortunately, a naive adaptation of definitions from other domains, e.g., the image domain, is flawed \cite{la2021king}. 

In deep learning, robustness is often measured in terms of how much a bounded perturbation (e.g., with respect to an $\ell_p$-norm) of a test set input affects the output of a network \cite{szegedy2013intriguing}. 
For NLP, bounded $\ell_p$-norm perturbations applied directly to an input do not preserve its semantic meaning or syntactic structure and are therefore linguistically uninteresting. Further, $\ell_p$ distance measures in the embedding space do not reflect %
how the input perturbation has affected the representation of key linguistic features, which are extracted by the language model from the input data.     

In this paper, %
we propose a framework to evaluate the syntactic consistency and robustness of linguistic representations that leverages probing tasks \cite{conneau2018you, manning2020emergent}, namely, neural networks trained directly on the representation embedding to evaluate the representation's ability to encode a specific linguistic phenomenon, such as the syntax tree of a sentence. 
To this end, we propose an efficient probing method to perturb the input text so that its syntax (or context) is largely preserved. We validate the perturbations to show that they can serve as an effective proxy of syntax-preserving perturbations.
We focus on syntactic robustness, which informs our selection of probing tasks, but note that other tasks can be easily incorporated. To assess robustness, we aim to measure the performance of a language model to probing tasks on the original and perturbed datasets. More specifically, we define a measure of robustness in terms of aggregating (averaging) the worst-case drop of performance of a collection of probing tasks over a given dataset, for a given perturbation budget, which then captures the model's ability to encode the linguistic phenomena, and is therefore more appropriate for NLP settings.

In principle, our methodology for evaluating robustness of linguistic representations allows us to benchmark LLMs and can be used by others to guide the development of models that optimize for robust syntactic understanding, which we find to be universally lacking. In addition, we demonstrate the ability of the proposed metrics to offer novel insights and perspectives into the workings of LLMs. In particular, we show that, despite conventional wisdom, context-dependent LLMs (BERT) are just as syntactically brittle as context-free embeddings (Word2Vec), and that deeper latent features provide as much syntactic robustness as shallow features. We also offer a critique of the effect of fine-tuning of such representations. 
We choose 6 datasets from the English Universal Dependencies \cite{nivre2016universal}, which are representative of different linguistic registers, and perform experiments on both standard embeddings (Wod2Vec and GloVe) and modern LLMs (BERT and RoBERTa)  \cite{devlin2018bert,goyal2021exposing,liu2019roberta,mikolov2013distributed} on %
4 representative syntactic probing tasks from \cite{manning2020emergent}, i.e., structural probe \cite{hewitt2019structural}, part-of-speech (POS) tagging, root identification and calculation of the depth of a sentence's syntax tree. We use two complementary sources of perturbations.
The former method is grounded in the utilization of WordNet synonyms~\cite{miller1998wordnet}, which we subsequently augment with constraints designed to uphold the syntactic structure of a sentence while modifying the maximum number of words permissible within a predetermined perturbation threshold. Conversely, the latter approach is centered around word prediction and relies on GPT-2~\cite{radford2019language}.
WordNet facilitates the selection of perturbations that maintain syntactic integrity, whereas GPT-2, along with other language models, produces substitutions that typically do not retain the original sentence's syntactic coherence. However, they compensate for this limitation by demonstrating a heightened awareness of contextual factors.

In summary, in this work we make the following contributions: %

\begin{itemize}
  \item Propose measures to evaluate robustness of linguistic representations that leverage probing tasks.
  \item Develop a methodology for analyzing an LLM's ability to robustly capture complex syntactical information underlying its training data.
  \item Demonstrate how our robustness metrics reveal that context-free representations are 
  equally brittle to manipulations as more sophisticated context-dependent representations.
  \item Provide empirically insightful observations into feature collapse, training duration, and depth of pre-trained LLM heads from the robustness perspective.
\end{itemize}

In addition to these empirical observations, we draw attention to
the brittleness of emergent syntactic representations of language models as %
a contribution to the debate about the capabilities of LLMs. The code, trained models and logs are made available for reproducibility.\footnote{The code to replicate the experiments of this paper is available at the following repository: \url{https://github.com/EmanueleLM/emergent-linguistic-structures}.}

\section{Related Works}
In this section, we first overview the linguistic models we study. Next,
we discuss recent methods aimed at extracting the syntactic structure represented by a language model. Finally, we summarise a series of works that revealed weaknesses in language understanding captured by these models. 

\paragraph{Linguistic representations} 

Early attempts to represent language were in the form of bag-of-words or binary/one-hot encodings~\cite{manning1999foundations}. The success of deep learning led to the increasing reliance on vector representations of language (word embeddings) in NLP tasks
\cite{turian2010word}. Word embeddings such as Word2Vec\cite{mikolov2013distributed} and GloVe\cite{pennington2014glove} translate one-hot encoded words and embed them into real-valued vectors such that similar words are mapped to similar vectors. %
In the past decade, researchers developed linguistic representations whose symbols are independent from each other: we call such representation context-free~\cite{manning1999foundations, mikolov2013distributed, pennington2014glove}. Only recently, with the improvement of training procedures and the capability of deep learning models to `digest' massive datasets, representations where each  symbol depends on the context in which it appears became possible~\cite{peters-etal-2018-deep}, thus better embodying the distributional hypothesis~\cite{rubenstein1965contextual}. We refer to this approach  as context-dependent, e.g., Large Language Models (LLM)~\cite{devlin2018bert,peters-etal-2018-deep}. %

\paragraph{Extracting syntactic structures}
Many works have investigated  %
whether representations embed the structure of a language, with a particular focus on LLMs~\cite{goldberg2019assessing,jawahar2019does,manning2020emergent} recently, and context-free representations~\cite{limisiewicz2020syntax} prior to this.
There is an ongoing debate
on whether representations can embody complex syntactic structures~\cite{hessel2021effective,sinha2021masked}.
Studies include 
assessing grammaticality of a representation~\cite{marvin2018targeted}
and
extracting grammars from representations, with works ranging from linguistics~\cite{dunn2017computational,dunn2021learned} to formal languages~\cite{shen2018straight} and NLP~\cite{kim2020pre,marevcek2019balustrades}. 

\paragraph{Alignment in NLP}
The impressive performance of modern LLMs has led to claims that they have ``mastered'' language \cite{johnson_iziev_2022}.
This claim has been disputed by a series of %
works seeking to contextualize the results of LLMs, in particular showing their lack of ``natural language understanding'' \cite{bender2020climbing}. In~\cite{bolukbasi2016man} the authors show that context-free representations can be gender biased%
. Considerable biases are also found to exist in context-dependent representations in \cite{le2020adversarial}. In addition to bias, \cite{bender2021dangers} highlights the multitude of ways in which LLMs are not well aligned with stakeholder values. %
In \cite{niven2019probing}, the authors highlight the language models' failure to perform basic linguistic reasoning tasks, while in \cite{floridi2020gpt} their scope limitations.  %

The works closest to our paper are those that study the robustness of NLP models. While human understanding of linguistic structures is very robust~\cite{gibson2019efficiency}, the robustness of NLP models is still far from being achieved~\cite{wallace2019universal}, as prominent works over-focus on a notion of adversarial robustness~\cite{huang2019achieving,jia2019certified,la2020assessing} that is linguistically flawed~\cite{la2021king,xu2020elephant}.  %
Practically speaking, %
robustness is measured and guaranteed either in the embedding space, hence w.r.t. bounded $\ell_p$-norm changes of a sentence's embedding representation \cite{huang2019achieving,la2021guaranteed}, or through discrete, semantically enhanced replacements~\cite{alzantot2018generating,dong2021towards,ribeiro2020beyond,la2021king}, which do not capture linguistic structures such as syntax.

\begin{figure*}
     \centering
     \includegraphics[width=1.0\linewidth]{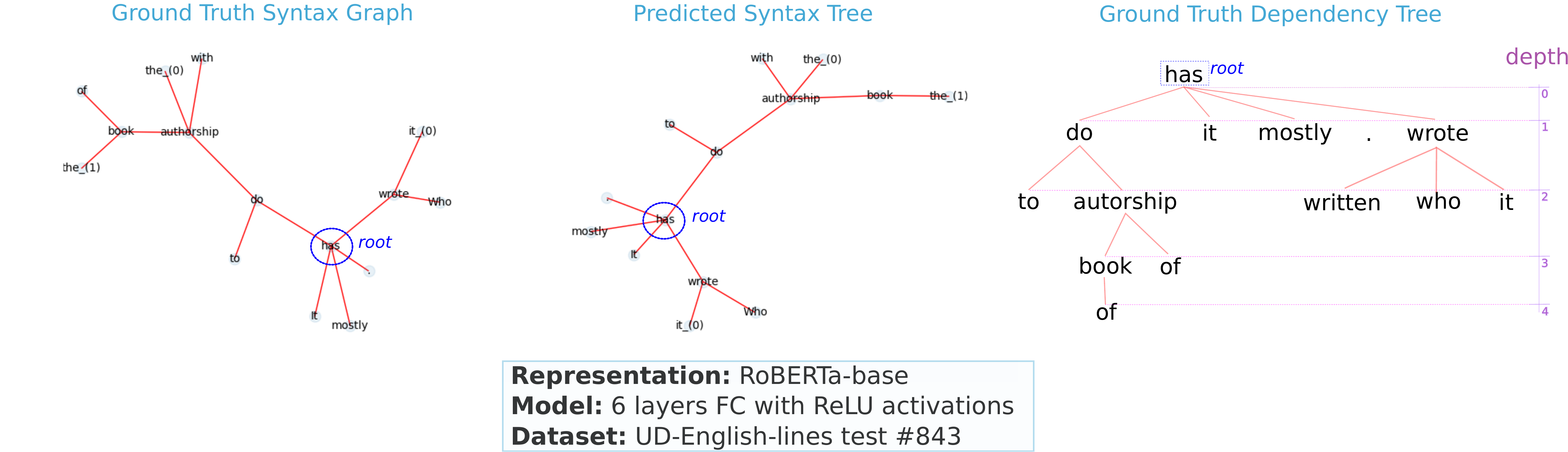}
     \caption{%
     A syntax graph reconstructed via the structural probe task from a RoBERTa representation is shown in the middle; for comparison, the ground truth structure is sketched on the left. On the right, the same structure is displayed as a dependency tree  (annotated with additional information so that dependencies and hierarchies between words are made clear) so that other supervised tasks can be instantiated, e.g., identifying the \textcolor{blue}{root}, or computing the \textcolor{realpurple}{depth} of the tree.}
     \label{fig:roberta-p1-pud-english-lines}
\end{figure*}
\section{Background and Notation}
In this %
section, we present the notation and concepts that we use to frame our methodology. A sentence $s=\{s_1, .., s_l\}$ is a finite sequence of $l>0$ symbols (here words) defined over finite vocabulary $\Sigma$. %
A sense of grammaticality is given by linguistic rules. A  linguistic rule assesses the violation of a property by a sentence $s$ and we denote a sentence $s$ satisfying a rule, $\mathbf{R}$, with %
$\mathbf{R} \models s$. %
A language, %
$\mathbf{L}$, is defined by an alphabet $\Sigma$ and a (possibly infinite) set of rules $\mathbf{R}=\{{R}_1, .., {R}_n\}$.

The application of neural networks to language became possible thanks to a numerical representation of sentences~\cite{mikolov2013distributed}. Given a sentence $s$ comprising $l>0$ symbols  from a language 
$\mathbf{L}$, a linguistic representation $\psi^{\theta}$, where $\theta$ are parameters, maps $s$ into a $(l \cdot d)$-dimensional vector of real numbers, i.e., 
$\psi^{\theta}: s \in \mathbf{L} \xrightarrow{} \mathbb{R}^{l \cdot d}$.
A linguistic representation $\psi^{\theta}$ is said to be context-free (or independent from the context) when the representation of each word is independent from the other words, i.e., $\psi^{\theta}(s_i | s \setminus s_i)=\psi^{\theta}(s_i)$. 
Otherwise, it is said 
to be context-dependent (or dependent on the context).

\section{Methodology}
Given a linguistic embedding $\psi^{\theta}$ and a sentence $s$, the central question of interest in this work is what information does $\psi^{\theta}$ extract robustly from $s$? To answer this question we consider using perturbation-based analysis. Specifically, given another sentence $s'$ that is %
similar to $s$, how does $\psi^{\theta}(s)$ differ from $\psi^{\theta}(s')$? While such perturbation analysis is reminiscent of adversarial robustness in the image domain~\cite{szegedy2013intriguing}, we highlight that a naive adaptation to NLP is devoid of the nuance of natural language and inappropriate for this setting \cite{la2021king}. %
We address this shortcoming with a two-phase framework. Firstly, we seek to gain insights into the syntactic properties understood by a language model through the use of probing tasks. Secondly, we propose an efficient scheme for computing perturbations that aim to preserve the sentence's original syntax, and study how such perturbations affect the model insights from the probing task.

\subsection{Probing Tasks for Model Introspection}
 
Recently, probing tasks have been introduced as a linguistically relevant measure of a model's understanding of complex linguistic phenomena, often grouped into surface, syntax, and semantic probing tasks~\cite{conneau2018you}. %
A probing task is a simple, non-challenging task used to extract linguistically meaningful information about a single facet of a language model, e.g., the subject number task requires us to extract the number of subjects in a sentence from its embedding. 
Probing tasks are classifiers trained directly on the representation
embedding to evaluate the representation’s ability to encode a specific linguistic
phenomenon.
The key idea here is that the probing task be linguistically specific -- testing the representation of a specific phenomenon -- and simple enough that strong performance on the task indicates, without bias, that a language model has accurately represented the given linguistic phenomenon. 
We select four probing tasks, %
which we design 
to assess the presence of syntactic structures in linguistic representations, but stress that our framework could be extended to any of the ten probing tasks presented in \cite{conneau2018you}.

We first define a generic probing task, which serves as a basis to describe the four syntactic tasks that feature in our robustness framework.
\begin{definition} \label{def:probing-task}
(Probing Task) Given a set $S=\{s^{(1)}, .., s^{(n)}\}$ of $n>0$ sentences from a language $\mathbf{L}$, each paired with a label $T=\{t^{(1)}, .., t^{(n)}\}$, a \emph{probing task} consists of finding a mapping $f$ from each sentence representation $\psi^{\theta}(s)$ s.t. 
$\mathbb{E}_{(s^{(i)},t^{(i)})\sim (S, T)}\, [\mathcal{L}(f(\psi^{\theta}(s)), t)]>p$, where $\mathcal{L}$ is a measure of performance of such a reconstruction, and $p$ some positive quantity that certifies a given level of performance.
\end{definition}

The first syntactic probing task we propose to study is the \textit{syntax reconstruction} task. An accurate understanding of the information content of a sentence $s$ depends on the  reader's ability to understand the intra-word relationships in $s$. This is not just true for natural language, but also for programming languages where parse trees are important to understand source code. 
A syntax tree $t$ is an undirected, acyclic graph $G := (s, A)$, where the words of $s$ are vertices and $A$ is an edge list which contains an edge between two words if they modify each other or are contextually linked, see~\cite{manning1999foundations} for more details. There are two standard representations of syntax trees in NLP and linguistics, namely dependency and constituency trees. In the former, each word corresponds to a node and the tree structure reflects the word order, while, in the latter, words themselves are terminal nodes whose order follows the `bare phrase structure' (as per the minimalist program by Noam Chomsky~\cite{10.7551/mitpress/9780262527347.003.0003}). In this paper, we work with dependency tree representations, but the methodology and the results can be extended to the constituency representation standard.
Formally, the syntax reconstruction probing task is given as: 

\begin{definition} \label{def:structural-probe}
(Syntax Reconstruction) Given a set $S=\{s^{(1)}, .., s^{(n)}\}$ of $n>0$ sentences from a language $\mathbf{L}$ and their syntax-tree representation $T=\{t^{(1)}, .., t^{(n)}\}$, \emph{syntax reconstruction} is a probing task $f$ from $S$ to $T$ that guarantees sufficient performance.
\end{definition}

\begin{wrapfigure}{R}{0.5\textwidth}
\begin{minipage}{0.5\textwidth}
     \includegraphics[width=1\linewidth]{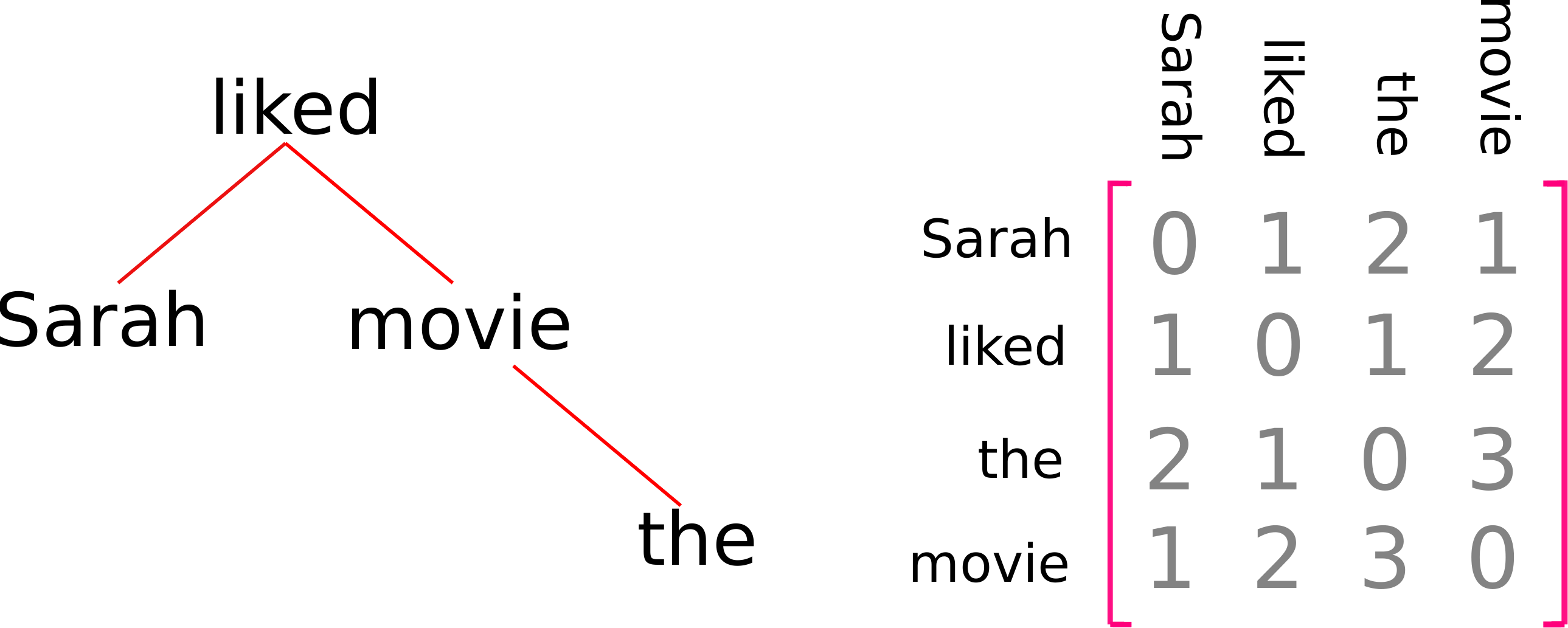}
     \caption{The dependency parse tree of a sentence (left), alongside the matrix of distances between pairs of words in the tree (right). 
     }
     \label{fig:probing-minipage}
\end{minipage}
\end{wrapfigure}

In practice, the syntax probing task consists of extracting, from a sentence representation, the distance between each pair of words, as they are arranged in the dependency parse tree of the sentence itself (see Figure~\ref{fig:probing-minipage}): the task
is commonly used as a proxy of the capabilities of a
representation to recognize the mutual dependency relationships between words in a sentence, represented as a directed graph. 
Probes are usually linear~\cite{manning2020emergent}, as one wants to assess how representations encode features that are immediately available to solve the task~\cite{niven2019probing}, though there has been recent criticism of the excessive simplicity of linear probes compared to non-linear ones \cite{pimentel2020information,white2021non}.
\par
Using probing tasks to assess the capabilities of a model has become a popular approach with the development of increasingly complex linguistic representations. However, some studies have shown that probes can only reveal the correlation between the traces of a symbolic structure in a representation and its performance on a task~\cite{belinkov2022probing,ravichander2020probing}. In our work, we use probes to provide evidence of the existence of syntactic structures in linguistic representations, rather than testing their performance on higher-level NLP tasks.
We show an example of a dependency syntax tree and its reconstruction in Figure \ref{fig:roberta-p1-pud-english-lines}.

The second probing task disregards intra-word relationships and focuses on a language model's ability to identify the part of speech of a given word. Formally, the \textit{part-of-speech (POS) tagging} task is given as:

\begin{definition} \label{def:pos-task}
(POS-tagging) Given a set $S=\{s^{(1)}, .., s^{(n)}\}$ of $n>0$ sentences from a language $\mathbf{L}$ and 
the POS-tags for each sentence, 
$POS=\{pos^{(1)}, .., pos^{(n)}\}$, \emph{POS-tag reconstruction}
is a probing task $g$ from $S$ to $POS$ that guarantees sufficient performance.
\end{definition}

\begin{figure*}
     \centering
     \includegraphics[width=0.8\linewidth]{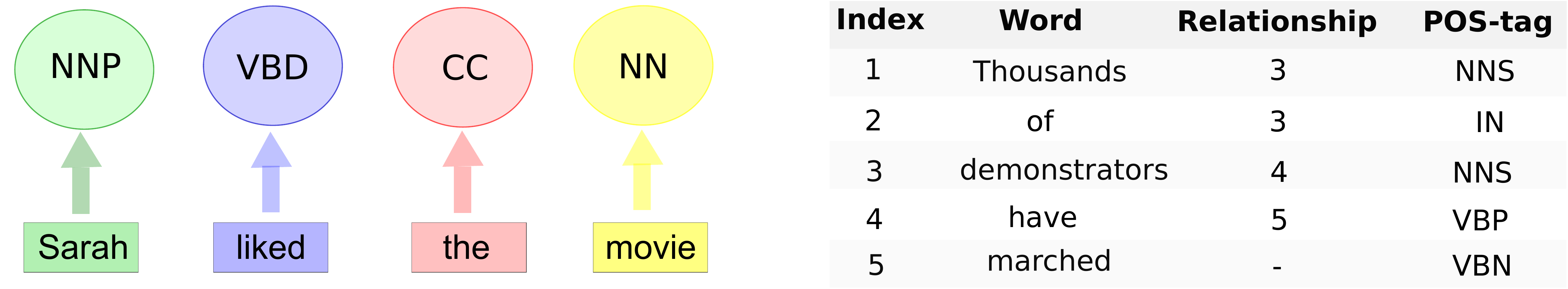}
     \caption{A sentence with its POS tags (left). A sentence in the CONLL format, used to test a model on multiple syntactic tasks (right).
     }
     \label{fig:pos-tag-example}
\end{figure*}

This task is commonly used as a proxy of the capabilities of a 
representation to represent the role of a word in its context: an example is shown in Figure \ref{fig:pos-tag-example}. In conjunction, these two tasks allow us to inspect how a language model identifies and semantically links entities in a sentence, thus giving us a comprehensive, linguistically-informed perspective on what is captured by a language model. %

We complete the benchmark with two further syntactic tasks, namely root identification and the tree-depth estimation, which we present below.

\begin{definition} \label{def:root-task}
(Root Identification) 
Given $S$ and $T$ as in Def.~\ref{def:structural-probe},
and the root of the tree 
$R=\{r^{(1)}, .., r^{(n)}\} \ \text{where} \ r^{(i)} \in t^{(i)}$, \emph{root identification} is a probing task $h$ from $S$ to $R$ that guarantees sufficient performance.
\end{definition}

\begin{definition} \label{def:depth-task}
(Tree-depth Estimation) 
Given $S$ and $T$ as in Def.~\ref{def:structural-probe},
and the depth of the tree 
$D=\{d^{(1)}, .., d^{(n)}\} \ \text{where} \ d^{(i)} \in \mathbb{N}^+$, \emph{tree-depth estimation} is a probing task $u$ from $S$ to $D$ that guarantees sufficient performance.
\end{definition}

With tasks in Def.~\ref{def:root-task} and~\ref{def:depth-task}, we assess a representation's capacity to distil single units of information (root and depth), %
which can be extracted from a tree's sentence representation. We sketch the two tasks in Figure~\ref{fig:roberta-p1-pud-english-lines} (right). When compared to the structural probe task, root identification and tree-depth are easier to solve: in fact, they are meant to show to what extent low-order syntax information, as opposed to high-order encoded by structural probe, is encoded in a linguistic representation.

\subsection{Syntax-Preserving Perturbation Analysis}

\begin{figure*}
     \centering
     \includegraphics[width=1.0\linewidth]{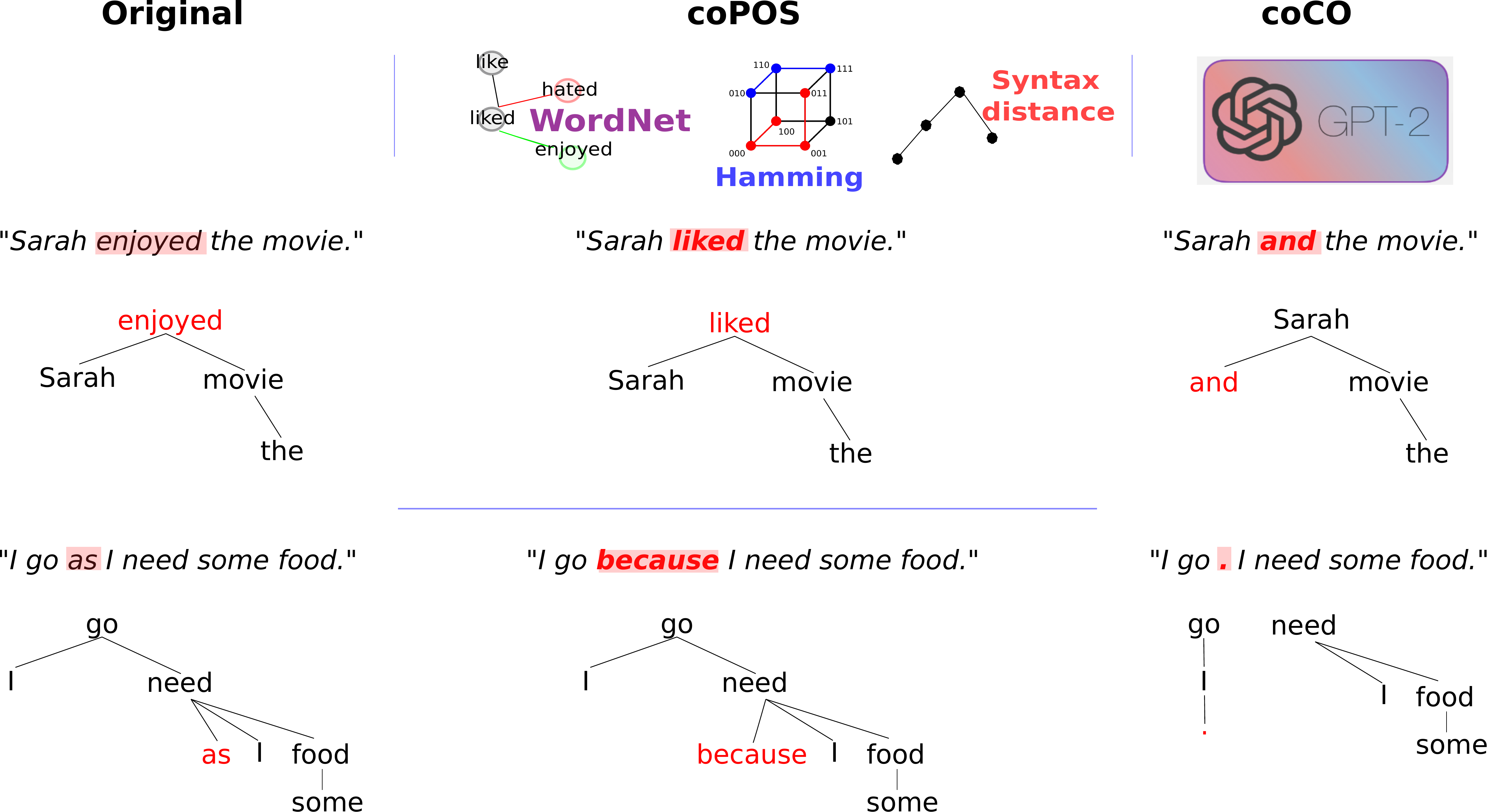}
     \caption{Two examples of coPOS and coCO perturbations applied on clean input texts, and the resulting syntax trees induced by such alterations. Words perturbed are highlighted in \textcolor{red}{red}. coPOS perturbations are designed to minimize the probability of disrupting the syntax of a sentence (such as the substitution of \texttt{`as'} with \texttt{`because'}), %
     while coCO can possibly disrupt it (e.g., substitution of \texttt{`and'} with a period). 
     }
     \label{fig:example-copos-coco}
\end{figure*}

The second phase of our methodology involves perturbation-based analysis. It is widely known and %
confirmed by neuroscience that human language exhibits very robust linguistic representations~\cite{chomsky2009syntactic,gibson2019efficiency}, while NLP models suffer from brittleness against perturbations, which are often easily transferable across models yet difficult to detect~\cite{kuleshov2018adversarial}. Though many works have shown how brittle NLP models are in the presence of bounded attacks on embedding space \cite{la2020assessing}, such attacks do not necessarily preserve human meaning and are therefore arguably of questionable merit \cite{la2021king}.  We define two types of perturbations: the first aims to preserve syntax (referred to as coPOS) and constitutes the backbone of our empirical evaluation; the second exploits context to preserve the semantics (coCO), and is introduced to strengthen our comparison of models' syntactic robustness. We further add, as baseline, a perturbation method with words randomly sampled from the English vocabulary. %
We now introduce 
the coPOS and the coCO perturbation methods, which are illustrated in Figure~\ref{fig:example-copos-coco}.

\begin{definition} \label{def:copos}
(Consistent POS Substitution) A \emph{consistent POS substitution (coPOS)} consists of the replacement of one or more words in a sentence $s$ with words that keep unaltered the POS-tag of the perturbed sentence, i.e., if $s$/$s'$ are the original/perturbed sentence, $pos$/$pos'$ the ground-truth POS-tag of $s$/$s'$, and $s'=sub(s)$, the perturbation procedure, then it holds for coPOS that $sub(s) \implies pos \equiv pos'$.
\end{definition}

Ensuring that a perturbation satisfies the coPOS definition enables interpretability of our results. Specifically, a coPOS perturbation 
is built with the intent to preserve the word's  syntactic role in a sentence, and therefore one can impute any probe misclassifications %
to a lack of robustness of the linguistic representation. Since guaranteeing that a perturbation always preserves its coPOS tag is challenging due to the intrinsic complexity of natural language, we rely on an efficient algorithmic implementation to generate proxy coPOS perturbations, described in Section~\ref{algorithms}, which we carefully validate on the datasets used in our experimental evaluation (see Section~\ref{sec:valid-pert}). %

\begin{definition} \label{def:coco}
(Context Consistent Substitution) A \emph{context consistent substitution (coCO)} consists of the replacement of one or more words in a sentence $s$ with a generative model that maintains semantic closeness but
does not strictly enforce the substitution to be syntactically coherent. While many alternative methods exist in the literature to generate coCo perturbations, we rely on GPT-2~\cite{radford2019language} next word predictions, which serves as a benchmark for syntactically-informed methods such as coPOS.
In other words, a substitution $w'$ of a word $w \in s$ is generated by a generative model $\phi$ 
conditioned on the context where the word appears, i.e., $w' = argmax_{w \in V}\phi(s | s \setminus w)$.
\end{definition}

Below, we formally define the conditions under which we consider a linguistic model robust: informally, for a linguistic representation to be robust we desire it to accurately solve a family of probing tasks and behave consistently on slight syntax-preserving perturbations of an input text. We assume that coPOS substitutions are used as perturbations, %
but note that the concept of linguistic robustness can also be instantiated with Def.~\ref{def:coco}.

First, we introduce 
the notion of consistency of representations, termed $\epsilon$-robustness.
\begin{definition} \label{def:eps-dist}
($\epsilon$-robust Representations) Given a linguistic representation $\psi^{\theta}$, a set of sentences $S$, a set of perturbed sentences $S'$ which are coPOS perturbations of $S$, and a measure of distance $dist: (s, s') \xrightarrow{} \mathbb{R}$ between representations (e.g., $\ell_p$-norm, cosine similarity), we say that the representation $\psi^{\theta}$ is \emph{$\epsilon$-robust w.r.t. $dist$} if 
$\ \forall (s,s') \in (S, S'), \ max (dist(\psi^{\theta}(s), \psi^{\theta}(s')))<\epsilon$. 
\end{definition}

Despite its simplicity, $\epsilon$-robustness is linguistically informed, as all sentences in $S'$ are coPOS to those in $S$, and thus we can be confident that the perturbations are syntactically consistent for the given probing task. 
Moreover, this metric can serve as a useful tool for developing robust language models, in the sense of maximizing $\epsilon$ while maintaining good performance on the underlying task.

While $\epsilon$-robust representations are desirable, what is more informative is the ability for a representation, $\psi^{\theta}$, to be robust not just with respect to a distance metric, but with respect to a probing task. Formally, we define a language model $\psi^{\theta}$ to be 
\textit{syntactically robust} if
the performance on multiple proxy tasks is not adversely affected by perturbations that are close in some representation space (e.g., Def.~\ref{def:copos}).

\begin{definition} \label{def:robust-ling-rep}
(Syntactically Robust Representation) Given an input $s$, its representation $\psi^{\theta}(s)$, a set of probing tasks $\{\mathbf{T_1}, .., \mathbf{T_m}\}$, a set of mappings $\{f_1(s), .., f_m(s)\}$ that take as input the representation $\psi^{\theta}(s)$ and solve the respective i-th probing task, a set of strictly positive quantities $\{\tau_1, .., \tau_m\}$ and a small quantity $\epsilon>0$, a set of measures of performance on each task $\{\mathcal{L}_1, .., \mathcal{L}_m\}$, a %
consistent perturbation $s'=sub(s)$, and a measure of distance between representations $dist: (s, s') \xrightarrow{} \mathbb{R}$, $\psi^{\theta}$ is \emph{syntactically robust}
iff $\forall (\mathbf{T}_i, f_i, \mathcal{L}_i, \tau_i) \in (\mathbf{T}, f, \mathcal{L}, \tau), \ dist(\psi^{\theta}(s), \psi^{\theta}(s'))<\epsilon \implies \mathcal{L}_i(f_i(\psi^{\theta}(s)), f_i(\psi^{\theta}(s')))<\tau_i$.
\end{definition}

\begin{figure*}
     \centering
     \includegraphics[width=1.0\linewidth]{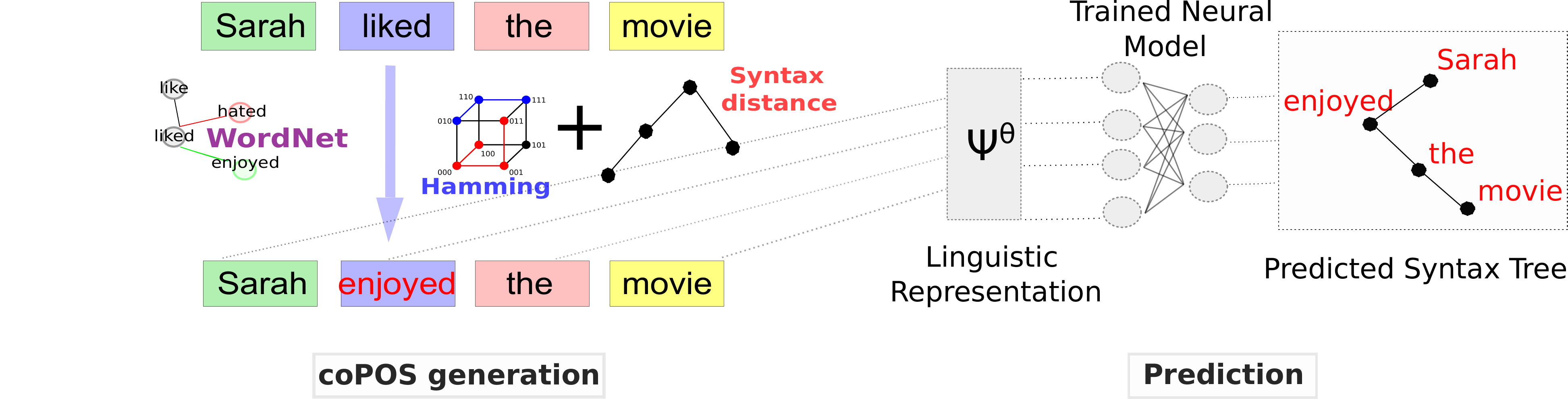}
     \caption{An example of a perturbed sentence $s'$ obtained through a coPOS perturbation. 
     Candidate substitutions are sampled from a pool of WordNet synonyms, from which we select the one that maximizes the Hamming distance and minimizes the syntactic disruption w.r.t. the original input, see Section~\ref{algorithms} for details. The perturbation is then fed, through a linguistic representation $\psi^{\theta}$, to %
     a probe (neural network trained directly on the representation)  that in turn predicts its syntax tree. 
     }
     \label{fig:running-example}
\end{figure*}

\section{Algorithm for Evaluating Robustness}\label{algorithms}

In this section, we describe a procedure to assess the robustness of the syntactic structures encoded by a linguistic representation%
, as formalized in Def. \ref{def:robust-ling-rep}.  We outline the full algorithm and give step-by-step comments 
in the Appendix, Section~\ref{app:algorithm}. 

The general framework takes as input a language model $\psi^{\theta}$, a set of 
$m$ probing tasks $\{\mathbf{T}_i\}_{i=1}^{m}$, performance metrics for each task $\{\mathcal{L}_i\}_{i=1}^{m}$, a perturbation function \textit{sub}, and two constants $\tau$ and $k$. For each task $\mathbf{T}$, we sample $n$ sentences and for each sentence $s$ we compute $k$ coPOS (or coCO, alternatively) perturbations $s'$, where each coPOS perturbation modifies $\tau$-many words. We then use the performance measure corresponding to the task to measure $\mathcal{L}(f(\psi^{\theta}(s)), f(\psi^{\theta}(s')))$ and take the perturbed sentence $s'$ that maximizes this quantity to be the %
approximate (since we sample finitely many replacements only) worst-case perturbation. Then we record the drop in performance that $s'$ causes. Finally, we take the average drop in performance across all $n$ sentences to be an approximate measure of worst-case performance for the language model on the given probing task.

\paragraph{coPOS perturbations}\looseness=-1
Given an %
$l>0$ word long sentence $s$, we formulate a method to obtain a perturbed sentence $s'$, where $\tau \le l$ words in $s$ are replaced whilst keeping the syntax of the original input largely preserved.\footnote{WordNet synonyms, or any similar technique, are specifically crafted to maintain the syntactic structure of word replacements. However, it is important to acknowledge that no technique can offer an absolute guarantee of preserving syntactic integrity when replacing a word in a sentence with a generic alternative.} Our procedure is sketched in Algorithm~\ref{alg:copos}. 

\begin{wrapfigure}{R}{0.5\textwidth}
\begin{minipage}{0.5\textwidth}
\begin{algorithm}[H]
\caption{coPOS perturbations.}\label{alg:copos}
\begin{algorithmic}[1]
\Require $s, b, \tau, \text{WordNet}(\cdot, \cdot), \text{dist}_{\text{ham}}(\cdot, \cdot), \newline \text{dist}_{\text{syntax}}(\cdot, \cdot)$
\Ensure A coPOS perturbation.
\State $s^*, d_{h}^*, d_{t}^* \gets (s, 0., \text{inf})$ \label{alg:init}
\For{$j \in [1, .., b]$}
    \State $s' \gets \text{WordNet}(s, \tau)$\label{alg:wn} \Comment{Perturb $\tau$ random words in $s$ with synonyms}
    \State $d_h \gets \text{dist}_{\text{ham}}(s, s')$
    \State $d_t \gets \text{dist}_{\text{syntax}}(s, s')$
    \If {$d_{h} > d_{h}^* \wedge d_{t} < d_{t}^*$}\label{alg:comp}
    \State $s^*, d_{h}^*, d_{t}^*  \gets (s', d_{h}, d_{t})$
    \EndIf
\EndFor
\State \textbf{return} $s^*$
\end{algorithmic}
\end{algorithm}
\end{minipage}
\end{wrapfigure}

 We replace each candidate word in $s$ with one drawn from the WordNet synonym graph \cite{miller1998wordnet}. We further ensure that a perturbation is, among the input-perturbation pairs generated by a WordNet replacement, the one that minimizes the syntactic distance of the tree representations while maximizing the 
 Hamming distance between the actual sentences, i.e., the number of words that are actually perturbed. The syntactic distance of each pair of inputs and perturbations is computed via the Stanza dependency parser~\cite{qi2020stanza}, while the Hamming distance between two sentences is the number of word positions in which two words are different.\footnote{As two sentences in an input-perturbation pair have the same number of words, we do not need to rely on the Levenshtein distance.}
In practice, for each input, $b>0$ sentences are generated by perturbing $\tau$ words via WordNet (line~\ref{alg:wn}): the syntactic distance between the dependency tree of each input/perturbation pair is computed, alongside their Hamming distance (line~\ref{alg:comp}), which could be less than $\tau$ if WordNet does not return a viable substitution, and only the sentence that minimizes the syntactic distance while maximizing the Hamming is %
used to test the representation's robustness. 

While this procedure is designed to preserve syntax between $s$ and $s'$, the semantics in general is not: though one may want to introduce further constraints on the replacement procedure to ensure the semantics is preserved, our %
primary intent is to assess robustness against syntax manipulations. We will show in the experiments that, even for these simple proxy syntax-preserving perturbations, a linguistic representation's performance degrades %
sensibly and in some cases it is comparable with random guessing, which indicates that this perturbation scheme is powerful enough to benchmark current language models. Further, this method has a clear advantage in terms of simplicity and computational efficiency as multiple word substitutions can be parallelized. We sketch the aforementioned procedure in Figure \ref{fig:running-example} (left). 

\paragraph{coCO perturbations}\looseness=-1 
    Our results are complemented by experiments with %
    coCo perturbations (Def.~\ref{def:coco}), which consist of generating $\tau$ replacements via a conditioned LLM, as explained in Def.~\ref{def:coco}. While we employ GPT-2~\cite{radford2019language} for generating a replacement, any generative LLM, thus including masked LLMs such as BERT or RoBERTa~\cite{devlin2018bert}, %
    is suitable for this task. Further implementation details are provided %
    in the Appendix, while the process of coCO perturbation is sketched in Figure~\ref{fig:example-copos-coco}. 
    
\paragraph{Baseline perturbations}\looseness=-1
Finally, we add a baseline perturbation method that involves substituting $\tau>0$ words in a sentence with random replacements from the English vocabulary. In this case, the syntactic consistency of a sentence is not guaranteed to be maintained, and thus serves as a base case for our analysis. 

\subsection{Validating the Perturbation Methods}\label{sec:valid-pert}
In this section, we report the results of the validation process of the coPOS and the coCO perturbation methods. For each perturbation method, and for each dataset that we then employ in the experimental evaluation, we calculate the syntactic distance between a sentence syntax tree and a perturbed candidate: the distance between trees is automatically computed as the minimum number of operations of addition and deletion of a node to turn a tree into another, via Stanza~\cite{bauer-etal-2023-semgrex} dependency parses.
While in this work we report the results regarding the distance between dependency trees, as that is the representation provided by the CoNLL format, as well as that employed in~\cite{manning2020emergent}, our code permits to compute distance between sentences via their constituency representation.
\par
\begin{figure*}
     \centering
     \includegraphics[width=1\linewidth]{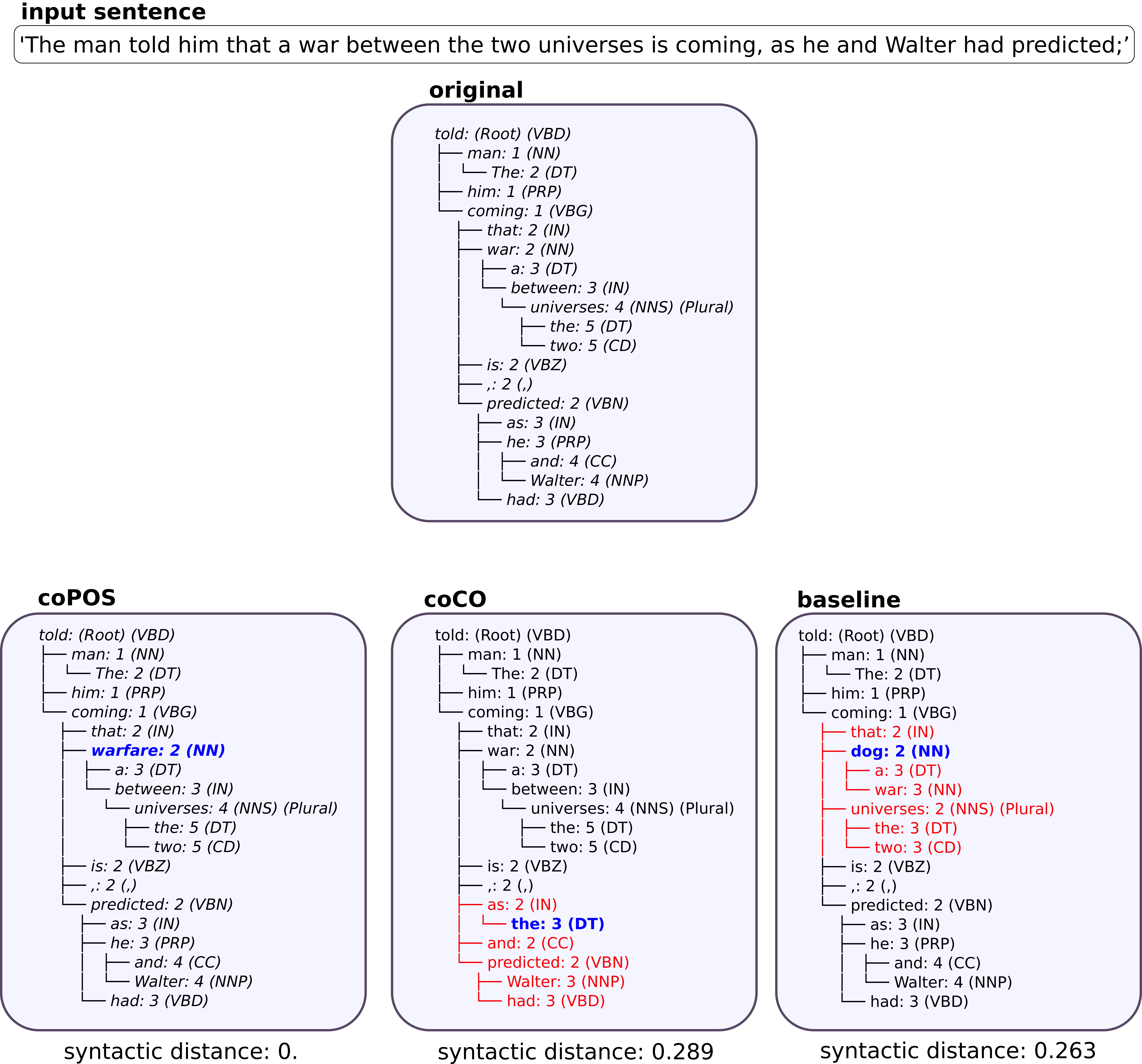}
     \caption{
     Comparison of the disruption induced on the 
     dependency syntax tree by different perturbation methods, along with the syntactic distance between trees. The representation of each dependency syntax tree has been compacted to make the effect of the perturbation methods clear, yet it is equivalent to that of Figures~\ref{fig:probing-minipage} and ~\ref{fig:example-copos-coco}. The example sentence belongs to the Ud-English-Pud dataset, and the perturbations are actual perturbations induced by our methods. In \textcolor{blue}{blue}, the single word that has been perturbed, while in \textcolor{red}{red} the perturbation induced by such perturbation on the tree. 
     }
     \label{fig:comparison-trees}
\end{figure*}
\begin{figure*}
     \centering
     \includegraphics[width=0.8\linewidth]{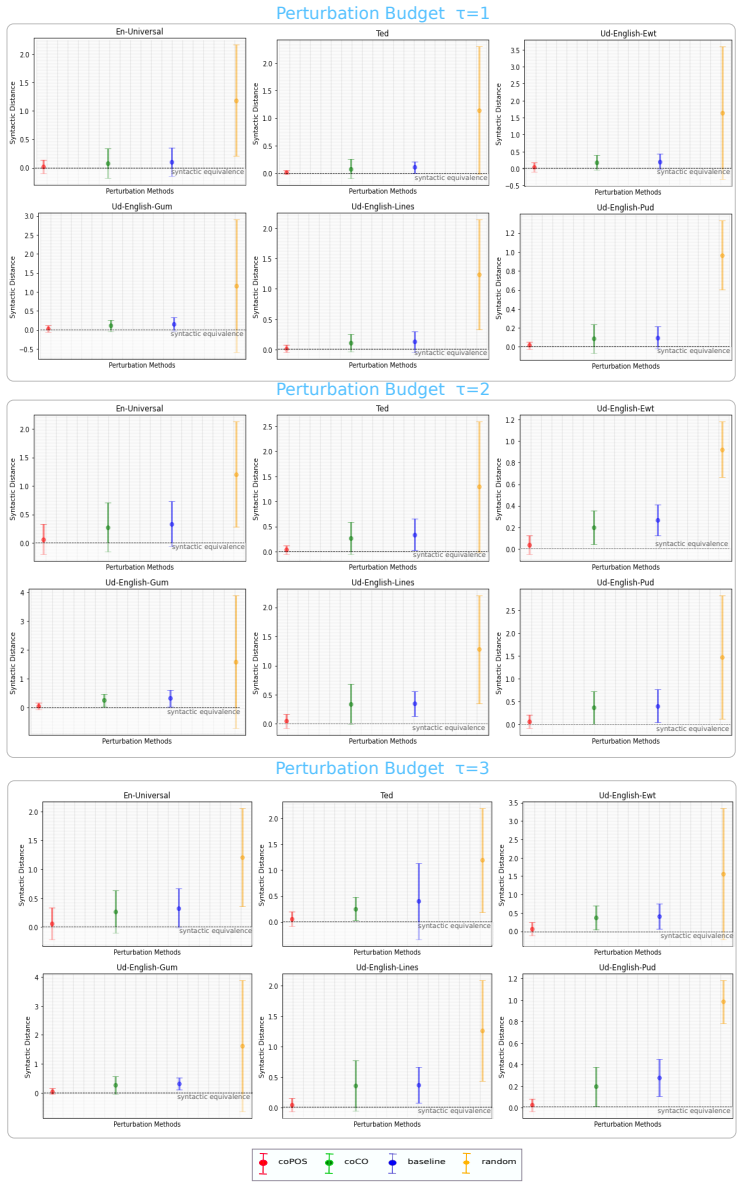}
     \caption{Tree distance, measured with Stanza, between an input and its perturbed version, 
     for different datasets and perturbation budgets: results are averaged over the entire dataset. The coPOS perturbation method (\textcolor{red}{red}) induces almost no disruption to a perturbation's syntax tree, being 
     always close to the level of syntactic equivalence, while injection of random words (\textcolor{blue}{blue}) and coPOS perturbations (\textcolor{greentxt}{green}) both induce some noticeable disruption. The disruption induced by comparing the syntax tree of two randomly picked up sentences that belong to the same dataset is reported for further comparison (\textcolor{orange}{orange}).
     }
     \label{fig:validate-mean-std}
\end{figure*}

Examples of the perturbed syntax tree of a sentence from the Ud-English-Pud dataset are shown in Figure~\ref{fig:comparison-trees} for each perturbation method. In Figure~\ref{fig:validate-mean-std}, we report, for each of the 6 dataset used in the experimental evaluation, the syntactic distance between trees, for the coPOS, coCO, and baseline method, with varying perturbation budget $\tau$ equal to 1, 2, and 3. We further compute the average distance between pairs of sentences randomly sampled from each dataset, and for which we expect the distance between trees to be higher than for any other method. %
\par
As one can notice from Figure~\ref{fig:validate-mean-std}, the coPOS method induces the least changes in a syntax tree, and it is thus expected to disrupt the performance of the probing tasks only if the representations are inherently brittle. On the other hand, both coCO and baseline are expected to challenge a probe's capacity to correctly represent a sentence's syntactic information. %
We will show with the proxy coPOS method that probes, and in turn their representations, are very brittle to syntax-preserving manipulations. 
\par
\begin{figure*}
     \centering
     \includegraphics[width=1\linewidth]{img/examples-copos-vs-coco-worst.pdf}
     \caption{
     Examples of sentences and worst-case coCO and coPOS perturbations that are reported in our experiments to highly disrupt the dependency syntax tree according to Stanza~\cite{qi2020stanza} (the syntactic distance between the original and perturnbed sentence is shown on the right). For each of the 6 CoNLL datasets, we show the original sentence on top. For coCO, perturbed words are highlighted in \textcolor{red}{red}) and replacements with empty words (which are allowed from the vocabulary) are denoted with a red rectangle \icon{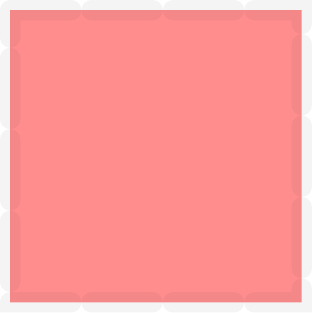}. For coPOS, perturbed words are highlighted in \textcolor{blue}{blue}. Results refer to the perturbation regime with $\tau=3$, i.e., where at most 3 words per-sentence are perturbed.
     }
     \label{fig:examples-copos-vs-coco-worst}
\end{figure*}

Finally, we show examples of perturbations that our methods produce. In Figure~\ref{fig:examples-copos-vs-coco-worst}, we report example sentences that induce, according to Stanza~\cite{qi2020stanza}, a high degree of disruption in the dependency syntax tree representation, with those produced by the coCO perturbation method the most disrupted for each input/perturbation pair (not counting the baseline, which almost surely disrupts the syntactic tree of a sentence through random replacements). On the other hand, the coPOS perturbation method can be seen to preserves the structure of each sentence. %
In Figure~\ref{fig:examples-copos-vs-coco}, we present examples of linguistically interesting perturbations, which do not induce the maximum syntactic disruption.

\begin{figure*}
     \centering
     \includegraphics[width=1\linewidth]{img/examples-copos-vs-coco.pdf}
     \caption{
     Examples of linguistically interesting sentences and perturbations, along with their syntactic distances (right) as calculated with Stanza~\cite{qi2020stanza}. For each of the 6 CoNLL datasets, we report the original sentence on top. For coCO, perturbed words are highlighted in \textcolor{red}{red}), while for coPOS in \textcolor{blue}{blue}. Results refer to the perturbation regime with $\tau=3$, i.e., where at most 3 words per-sentence are perturbed.
     }
     \label{fig:examples-copos-vs-coco}
\end{figure*}

\begin{figure*}
     \centering
     \includegraphics[width=1\linewidth]{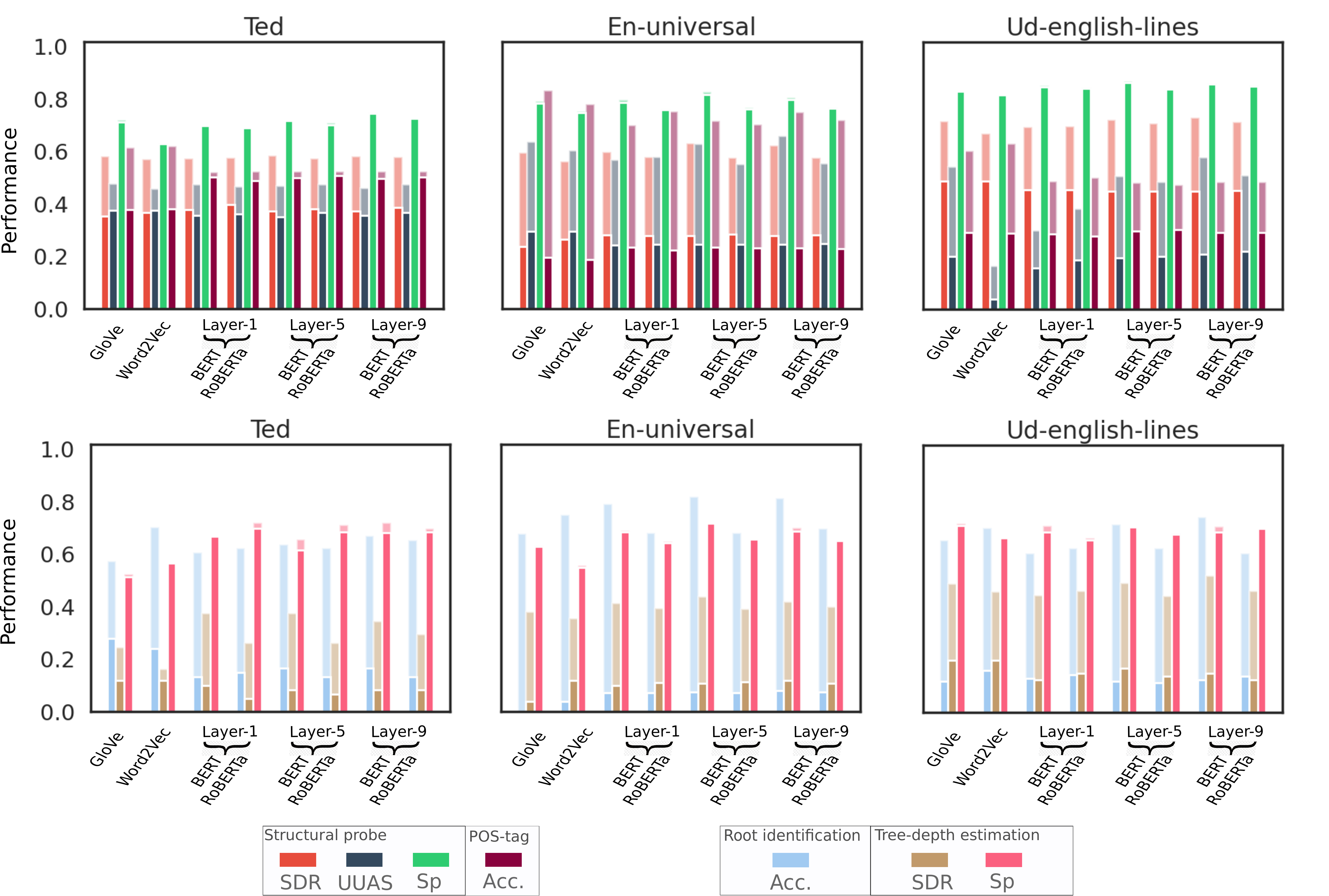}
     \caption{Performance of different linguistic representations on \textit{syntax reconstruction} and \textit{POS-tagging} probing tasks (top) and on \textit{root identification} (accuracy metric) and \textit{tree-depth estimation} (SDR and Spearman metric) probing tasks (bottom). For all plots, the performance of the probing tasks is reported as shaded bars, with the performance for the perturbed representation shown %
     as a solid overlapping bar: the results refer to the case where the coPOS perturbation budget $\tau$ is equal to 3 (i.e., at most 3 words per-sentence are perturbed). 
     Regardless of the embedding/representation employed, we observe severe brittleness of the syntactic representations. %
     }
     \label{fig:per-rob-t12-copos}
\end{figure*}

\section{Experiments}
We implement and empirically validate our framework by demonstrating how it can provide insights into the robustness of language models. We start with details on linguistic representations, datasets, and probing task models. We then discuss context-free and context-dependent linguistic representations, the effect of latent feature depth, and the duration of fine-tuning.\footnote{Further details on the implementation, the architectures, the training procedure and the configurations are provided in the Appendix. %
} Finally, we summarise the results of applying our framework in these settings.

\subsection{Experimental Setting}
\paragraph{Datasets and metrics}\looseness=-1
We assess the syntactic robustness of different language models using the probing tasks of \textit{syntax reconstruction}, \textit{POS-tagging}, \textit{root identification} and \textit{tree-depth estimation} 
on 6 datasets from the Universal Dependencies collection~\cite{nivre2016universal}. We chose datasets standardized according to the CONLL format~\cite{hajic2009conll}, which consists of sentences split into words, with each indexed and annotated with multiple syntactic information such as POS-tags and the relationship with other words/tokens. An example of a sentence in the CONLL format, with relationship tags between words and part-of-speech tags used to build the ground truth for our probing tasks, is given in  Figure~\ref{fig:pos-tag-example}.

We measure the performance on syntax reconstruction in terms of the `undirected unlabeled attachment score' (UUAS), i.e., the fraction of edges in the ground truth syntax tree that is correctly predicted by a model, the `same distance ratio' (SDR), i.e., the number of times a model correctly guesses the distance between each pair of words in the ground truth syntax tree, and the %
Spearman correlation (used in~\cite{manning2020emergent}), which summarizes the strength of the relationship between the matrix representation of the original vs. reconstructed syntax tree. With regards to POS-tagging, we evaluate a model in terms of the accuracy on estimating the correct POS-tag  as shown in Figure~\ref{fig:per-rob-t12-copos} (top). On root identification (see Def.~\ref{def:root-task}), we use the accuracy of correct vs. wrong estimates, while on tree-depth estimation (Def.~\ref{def:depth-task}) we use the SDR, along with the Spearman correlation, so as not to penalize %
models that do not infer the label exactly.

\begin{table*}[]\footnotesize
\centering
{{\textbf{\ref{tab:distance}.1 Robustness}}}
\vspace{1 mm}
\\
\begin{tabular}{|l|c|c|c|c|}
\hline
    \multirow{2}{2cm}{\textbf{}} & \multicolumn{3}{c|}{\textbf{Syntax Reconstruction}} & \textbf{POS-tagging}\\
    \cline{2-5}
    & \textbf{$\Delta$SDR} & \textbf{$\Delta$UUAS} & \textbf{$\Delta$Sp} & \textbf{$\Delta$Acc.}\\ \hline
\textbf{GloVe}             & $0.2066 \pm 0.1243$ & $0.2444\pm 0.1298$ & $0.0062\pm 0.0032$ &             $6.4897 \pm 3.0856$     \\ \hline
\textbf{Word2Vec}          & $0.1553 \pm 0.1161$ & $0.1145 \pm 0.1004$ & $0.0052 \pm 0.0048$ &             $6.831 \pm 2.2698$     \\ \hline
\textbf{BERT layer -1}     & $0.2011 \pm 0.0784$ & $0.1607 \pm 0.0835$ & $0.0032 \pm 0.0077$ &             $3.0803 \pm 2.9857$     \\ \hline
\textbf{RoBERTa layer -1}  & $0.193 \pm 0.0808$ & $0.1752 \pm 0.1117$ & $0.0019 \pm 0.0039$ &             $4.1293 \pm 3.3021$     \\ \hline
\textbf{BERT layer -5}     & $0.2287 \pm 0.0817$ & $0.2451 \pm 0.1212$ & $0.005 \pm 0.003$ &             $3.243 \pm 3.0814$     \\ \hline
\textbf{RoBERTa layer -5}  & $0.2038 \pm 0.0758$ & $0.2086 \pm 0.1052$ & $0.0024 \pm 0.0379$ &             $3.0607 \pm 3.0132$     \\ \hline
\textbf{BERT layer -9}     & $0.2307 \pm 0.0838$ & $0.281 \pm 0.1142$ & $0.0035 \pm 0.0037$ &             $3.544 \pm 3.2826$     \\ \hline
\textbf{RoBERTa layer -9}  & $0.2045 \pm 0.0763$ & $0.2148 \pm 0.1116$ & $0.0017 \pm 0.0023$ &             $ 3.4723\pm 3.1265$     \\ \hline
\end{tabular}
\vspace{2 mm}
\\
{{\textbf{\ref{tab:distance}.2 Robustness}}}
\\
\vspace{1 mm}
\begin{tabular}{|l|c|c|c|c|}
\hline
    \multirow{2}{2cm}{\textbf{}} &  \textbf{Root Identification} & \multicolumn{2}{c|}{\textbf{Tree Depth Estimation}}\\
    \cline{2-4}
    & \textbf{$\Delta$Acc.} & \textbf{$\Delta$Acc.} & \textbf{$\Delta$Sp}\\ \hline
\textbf{GloVe}             & $0.4387 \pm 0.1953$ &  $0.2663\pm 0.235$ & $0.0174\pm 0.0189$     \\ \hline
\textbf{Word2Vec}          & $0.5251 \pm 0.1123$ & $0.2582\pm 0.2712$ & $0.0093\pm 0.0285$     \\ \hline
\textbf{BERT layer -1}    & $0.5015 \pm 0.2135$ &  $0.3495\pm 0.2139$ & $0.0209\pm 0.0159$     \\ \hline
\textbf{RoBERTa layer -1} & $0.5286 \pm 0.1661$ &  $0.3193\pm 0.2348$ & $0.0261\pm 0.0126$     \\ \hline
\textbf{BERT layer -5}    & $0.6039 \pm 0.1383$ &  $0.3314\pm 0.2401$ & $0.0086\pm 0.0158$     \\ \hline
\textbf{RoBERTa layer -5} & $0.5211 \pm 0.168$ &  $0.2829\pm 0.2524$ & $-0.0062\pm 0.0379$     \\ \hline
\textbf{BERT layer -9}    & $0.612 \pm 0.1216$ & $0.3256\pm 0.2423$ & $0.0159\pm 0.018$     \\ \hline
\textbf{RoBERTa layer -9} & $0.5151 \pm 0.1773$ & $0.3312\pm 0.2365$ & $-0.0002\pm 0.0222$     \\ \hline
\end{tabular}
\vspace{2 mm}
\\
{{\textbf{\ref{tab:distance}.3 Distance/Similarity Metrics}}}
\\
\vspace{1 mm}
\begin{tabular}{|l|c|c|}
\hline
                  & \textbf{$\ell_2$-norm distance}  & \textbf{Cosine similarity}  \\ \hline
\textbf{GloVe}             & $0.023 \pm 0.0032$ & $0.9352 \pm 0.0132$     \\ \hline
\textbf{Word2Vec}           &  $0.0038 \pm 0.0005$ & $0.9231 \pm 0.0161$       \\ \hline
\textbf{BERT layer -1}    & $0.0299 \pm 0.0036$ & $0.8994 \pm 0.0221$       \\ \hline
\textbf{RoBERTa layer -1} & $0.0103 \pm 0.0013$ & $0.9835 \pm 0.0039$       \\ \hline
\textbf{BERT layer -5}    & $0.0388 \pm 0.0045$ & $0.926 \pm 0.0196$       \\ \hline
\textbf{RoBERTa layer -5} & $0.0259 \pm 0.0032$ & $0.9764 \pm 0.0059$       \\ \hline
\textbf{BERT layer -9}    & $0.0377 \pm 0.0045$ & $0.9296 \pm 0.0186$       \\ \hline
\textbf{RoBERTa layer -9} & $0.0188 \pm 0.0018$ & $0.9843 \pm 0.0026$       \\ \hline
\end{tabular}
\caption{%
Relationship between the syntactic robustness metrics for four probing tasks on coPOS perturbations with budget $\tau=2$ (top and middle row) and the distance between pairs of perturbed and original inputs measured via cosine similarity and $\ell_2$-norm distance (bottom row). %
The accuracy drop of the POS-tag task is reported as the number of words correctly guessed in both cases.
The reported standard deviation is measured by averaging over the 6 training corpora. Whilst the distance (similarity) between inputs and perturbations is low (high), we observe that all embeddings/representations are %
brittle to syntax-preserving perturbations. 
}
\label{tab:distance}
\end{table*}

\begin{table*}[]\footnotesize
\centering
{{\textbf{\ref{tab:distance-coCO}.1 Robustness}}}
\vspace{1 mm}
\\
\begin{tabular}{|l|c|c|c|c|}
\hline
    \multirow{2}{2cm}{\textbf{}} & \multicolumn{3}{c|}{\textbf{Syntax Reconstruction}} & \textbf{POS-tagging}\\
    \cline{2-5}
    & \textbf{$\Delta$SDR} & \textbf{$\Delta$UUAS} & \textbf{$\Delta$Sp} & \textbf{$\Delta$Acc.}\\ \hline
\textbf{GloVe}             & $0.2098 \pm 0.124$ & $0.2616 \pm 0.1411$ & $0.139 \pm 0.0106$ &             $6.743 \pm 3.0337$     \\ \hline
\textbf{Word2Vec}          & $0.1655 \pm 0.1114$ & $0.1232 \pm 0.1014$ & $0.0118 \pm 0.014$ &             $7.071 \pm 2.1687$     \\ \hline
\textbf{BERT layer -1}    & $0.208 \pm 0.0773$ & $0.1782 \pm 0.0862$ & $0.0285 \pm 0.0177$ &                 $3.1592 \pm 3.0359$      \\ \hline
\textbf{RoBERTa layer -1} & $0.1989 \pm 0.0823$ & $0.1951 \pm 0.116$ & $0.02 \pm 0.0146$ &                 $4.2887 \pm 3.3774$     \\ \hline
\textbf{BERT layer -5}    & $0.235 \pm 0.082$ & $0.267 \pm 0.1293$ & $0.0261 \pm 0.0191$ &                 $3.286 \pm 3.1258$      \\ \hline
\textbf{RoBERTa layer -5} & $0.2093 \pm 0.0767$ & $0.2319 \pm 0.1117$ & $0.0133 \pm 0.0126$ &                 $4.2887 \pm 3.3774$     \\ \hline
\textbf{BERT layer -9}    & $0.235 \pm 0.0859$ & $0.2988 \pm 0.154$ & $0.0162 \pm 0.0135$ &                 $3.613 \pm 3.3219$     \\ \hline

\textbf{RoBERTa layer -9} & $0.2109 \pm 0.0774$ & $0.2378 \pm 0.1227$ & $0.0135 \pm 0.012$ &                 $3.561 \pm 3.205$     \\ \hline
\end{tabular}
\vspace{2 mm}
\\
    {{\textbf{\ref{tab:distance-coCO}.2 Robustness}}}
    \\
    \vspace{1 mm}
\begin{tabular}{|l|c|c|c|c|}
\hline
    \multirow{2}{2cm}{\textbf{}} &  \textbf{Root Identification} & \multicolumn{2}{c|}{\textbf{Tree Depth Estimation}}\\
    \cline{2-4}
    & \textbf{$\Delta$Acc.} & \textbf{$\Delta$Acc.} & \textbf{$\Delta$Sp}\\ \hline
    \textbf{GloVe}             & $0.4987 \pm 0.1827$ &             $0.293, 0.2024$ & $0.0417, 0.0134$     \\ \hline
    \textbf{Word2Vec}      &    $0.5785 \pm 0.1462$ &             $0.2915, 0.2444$ & $0.0174, 0.0184$     \\ \hline
    
    \textbf{BERT layer -1}    &                 $0.5526 \pm 0.202$ &                 $0.3863, 0.2054$ & $0.0838, 0.0494$     \\ \hline
    \textbf{RoBERTa layer -1} & $0.5449 \pm 0.1804$ &                 $0.3567, 0.2166$ & $0.0993, 0.0531$     \\ \hline
    \textbf{BERT layer -5}    & $0.6374 \pm 0.1483$ &                 $0.3937, 0.2247$ & $0.0633, 0.0374$     \\ \hline
    \textbf{RoBERTa layer -5} & $0.5448 \pm 0.1735$ &                 $0.3267, 0.2338$ & $0.0672, 0.0215$     \\ \hline
    \textbf{BERT layer -9}    & $0.6293 \pm 0.1408$ &                 $0.3726, 0.2217$ & $0.054, 0.0512$     \\ \hline
    \textbf{RoBERTa layer -9} & $0.549 \pm 0.1786$ &                 $0.3613, 0.2317$ & $0.0471, 0.0326$     \\ \hline
    \end{tabular}
    \\
    \vspace{1 mm}
    {{\textbf{\ref{tab:distance-coCO}.3 Distance/Similarity Metrics}}}
    \\
    \vspace{1 mm}
    \begin{tabular}{|l|c|c|}
    \hline
                      & \textbf{$\ell_2$-norm distance}  & \textbf{Cosine similarity} \\ \hline
    \textbf{GloVe}             & $0.0344 \pm 0.0018$ & $0.8783 \pm 0.0271$   \\ \hline
    \textbf{Word2Vec}           & $0.0059 \pm 0.0002$ & $0.8572 \pm 0.0331$     \\ \hline
    \textbf{BERT layer -1}    & $0.0487 \pm 0.0063$ & $0.7951 \pm 0.0433$     \\ \hline
    \textbf{RoBERTa layer -1} & $0.0195 \pm 0.0016$ & $0.9597 \pm 0.0064$     \\ \hline
    \textbf{BERT layer -5}    & $0.0652 \pm 0.0058$ & $0.8432 \pm 0.0316$     \\ \hline
    \textbf{RoBERTa layer -5} & $0.0488 \pm 0.0037$ & $0.9416 \pm 0.0102$     \\ \hline
    \textbf{BERT layer -9}    & $0.058 \pm 0.004$ & $0.8768 \pm 0.0173$    \\ \hline
    \textbf{RoBERTa layer -9} & $0.0373 \pm 0.0024$ & $0.9557 \pm 0.0057$     \\ \hline
    \end{tabular}
\caption{%
Relationship between the syntactic robustness metrics  for four 
probing tasks on coCO perturbations with budget $\tau=2$ (top and middle row) and the distance between pairs of perturbed and original inputs measured via cosine similarity and $\ell_2$-norm distance (bottom row). %
The reported standard deviation is measured averaging over the 6 training corpora. The accuracy drop of the POS-tag task is reported as the number of words correctly guessed in both cases. Whilst the distance (similarity) between inputs and perturbations is low (high), we observe that all embeddings/representations are brittle to syntax-preserving perturbations.}
\label{tab:distance-coCO}
\end{table*}

\textit{Syntactic robustness} (Def.~\ref{def:robust-ling-rep}) is measured in terms of the drop in the performance of a model when targeted with a coPOS or a coCO perturbation, e.g., %
\text{$\Delta$UUAS} represents the drop of the UUAS metric on the syntax robustness task, comparing the performance on the original sentence with its perturbed version. Regarding POS-tags, we convert the drop
of accuracy into the more intuitive difference between words correctly guessed on the original sentence compared to its
perturbed version, denoted `\# Words Adv' in Table \ref{tab:distance}. 

\paragraph{Models and probing tasks}\looseness=-1 We perform our analyses on 4 linguistic representations, of which 2 are context-free, namely GloVe~\cite{pennington2014glove} and Word2Vec~\cite{mikolov2013distributed}, and 2 context-dependent, namely, BERT~\cite{devlin2018bert} and RoBERTa~\cite{liu2019roberta}. As LLMs employ deep attention-based architectures, we perform experiments on sentences distilled from the $-5^{th}$ (i.e., the fifth counting from the most external), the $-9^{th}$ and the last (i.e., $-1^{th}$ or the output) hidden layer of a representation. While in~\cite{manning2020emergent} it was observed that the most hidden layers are those that perform the best on syntactic tasks, we provide results also for an intermediate and the last hidden layer. %

For each probing task (in our setting, syntax reconstruction, POS-tagging, root
identification and tree-depth estimation, as per %
Def. \ref{def:structural-probe}, \ref{def:pos-task}, \ref{def:root-task} and \ref{def:depth-task}), %
we stack a deep neural network on top of a linguistic representation $\psi^{\theta}$, thus obtaining a set of models $\{f_1(s), .., f_m(s)\}$: we optimize each model $f_i$ via supervised learning on the $i$-th task $\mathbf{T_i}$, leaving the representation's parameters $\theta$ fixed.
We note that the measures of performance $\{\mathcal{L}_1, .., \mathcal{L}_m\}$ vary from a probing task to another, as we detail in the experimental evaluation.

When training the probing task models, we searched for a common architecture that achieves high performance for each of the four language models. We tested fully-connected, convolutional, and recurrent neural networks, and found that fully-connected (FC) probing models had the best performance across the language models.
For each combination of datasets, probing tasks, models, perturbation methods (coPOS, coCO), and for a varying perturbation budget $\tau$, we train a 3-layer deep FC network with a varying number of parameters in the order of 10M. In this sense, both the static and dynamic representations are kept fixed (i.e., their parameters are `frozen' at training time), not to invalidate the scope of the probing tasks and to allow full reproducibility of the results. This experimental evaluation accounts for approximately 900 distinct cases. %

\subsection{Empirical Evaluation of Syntactic Robustness}
We now report the results of our robustness evaluation; in particular, we can quantify the syntactic robustness of the representation $\psi^{\theta}$ according to Def.~\ref{def:robust-ling-rep}.

\paragraph{Performance on probing tasks}\looseness=-1 In Figure~\ref{fig:per-rob-t12-copos} we observe that, across the structural probe task and the POS-tag, all the models have similar performances, with an average POS-tag accuracy around 0.8 and syntax reconstruction SDR around 
0.7. Of the context-dependent language models, RoBERTa and BERT are comparable, though we cannot definitively conclude which one is better. Similarly, GloVe slightly outperforms Word2Vec on tree-depth estimation, while Word2Vec is better on the structural probe. Interestingly, we notice that word embeddings only slightly worse than BERT and RoBERTa. The same trends emerge on %
root identification and tree-depth estimation, as shown in Figure~\ref{fig:per-rob-t12-copos} (bottom): while BERT generally outperforms the competitors and Word2Vec struggles to compete, GloVe is a competitor of both the language models.
We conclude with a final remark on the %
lack of performance gap between GloVe and Word2Vec, as it suggests that 
pre-training a representation on local and global word co-occurrences \cite{pennington2014glove} does not help syntactic structures to emerge, a controversial yet intriguing discovery.\footnote{\url{http://languagelog.ldc.upenn.edu/myl/PinkerChomskyMIT.html}}

\begin{figure*}
     \centering
     \includegraphics[width=1.\linewidth]{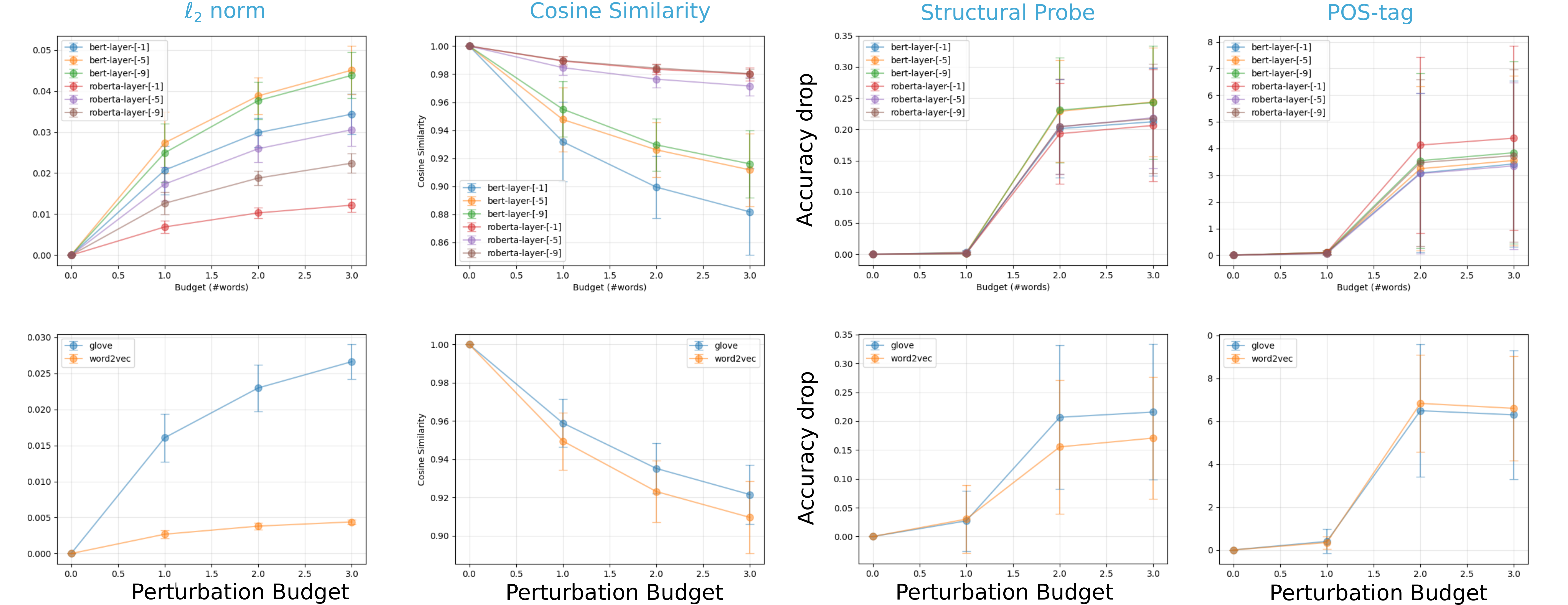}
     \caption{
     Left: For an increasing perturbation budget $\tau$ and the coPOS method, cosine similarity between perturbed and original sentences drops, while the $\ell_2$-norm increases. Right: It is clear that, even with $\tau=2$ (i.e., at most two words per-sentence are perturbed), the models' performance already experiences a significant drop (the higher the curve, the worse the model is on a syntactic task). Increasing the perturbation budget only slightly increases a drop of robustness. 
     }
     \label{fig:similarity-perfdrop}
\end{figure*}

\paragraph{Robustness on probing tasks}\looseness=-1
In %
Figure~\ref{fig:per-rob-t12-copos}, solid bars represent the performance of a trained model subjected to a coPOS perturbation under the perturbation budget of $\tau=3$: in other words, we generate an approximate worst-case coPOS perturbation given that we can only change at most 3 word in the given sentence. The drop in performance of a model on a task %
can be inferred via the gap between the solid and shaded bars (the latter corresponding to the unperturbed sentence). We notice that for each metric, excluding the Spearman correlation in %
tasks 1 and 4, discussed in the Appendix, there is a substantial drop in performance, which suggests that each language model represents a brittle understanding of syntactic concepts. While Spearman is used as a metric in~\cite{manning2020emergent}, in the Appendix, Section~\ref{app:spearman}, we discuss an interesting finding, validated by substantial empirical evidence, 
that we believe partially invalidates it as a proper metric to judge the syntactic similarity between trees extracted from linguistic representations.
In particular, in the syntax reconstruction task (Figure~\ref{fig:per-rob-t12-copos}, top) we notice a dramatic decrease in UUAS of more than $50\%$ on any task and any model, while the SDR drop is of around $20-30\%$. %
Thus, for each language model and dataset, the syntax reconstruction task is now %
incorrect more often than it is correct. For UUAS, this indicates that our coPOS scheme is able to find syntactically meaningful perturbations which reduce the performance of the model to random guessing. Secondly, we highlight that, for UUAS and SDR, the largest decrease in performance comes for the datasets for which the performance was the highest. This indicates that robust representation of syntax may be at odds with performance.
The same considerations are valid for the POS-tag probing task, with the accuracy that drops to a random guess with the perturbation budget $\tau$ equal to 1.
The coPOS perturbation method degrades the performance on root identification and tree-depth estimation as much as in the previous tasks (also in this case excluding the Spearman correlation, which we remind we discuss  %
in the Appendix). In fact, 
the performance drops to a random guess on any task and for any representation, which provides evidence that these representations are brittle, and thus not suited to cases of domain shifts.

\paragraph{On the correlation between robustness and sentence similarity}
In this batch of experiments, we keep track of the farthest distance between an input and its coPOS perturbation (see Def.~\ref{def:eps-dist}) using the $\ell_2$-norm and the cosine similarity between each pair of input/perturbation. We then measure the performance drop of each model to assess any correlation with the aforementioned measures of distance/similarity. As reported in Table~\ref{tab:distance}, we find that high drops in performance can be caused by perturbations with small $\ell_2$-norm compared to the unperturbed sentence, and conversely high cosine similarity. This confirms that linguistic representations are remarkably brittle even to local perturbations, i.e., those whose representation lies in the proximity of the original input. We replicate the results using the coCO perturbation method (Table~\ref{tab:distance-coCO}), which confirm that perturbations extracted via GPT conditioning are farther than the coPOS in the representation space, and equally effective at dismantling a model's robustness. Similar observations can be made with the baseline perturbation method, as reported in the Appendix, Table~\ref{tab:distance-tau2}.

\paragraph{Varying the perturbation budget}
While we have already shown that a small perturbation budget $\tau$ exposes a representation to effective performance-degrading attacks, we now investigate the relationship of the distance/similarity metrics and robustness with respect to a varying number of coPOS perturbations. In Figure~\ref{fig:similarity-perfdrop}, both the cosine similarity  and the $\ell_2$-norm behave monotonically as $\tau$ increases. %
Deeper LLM's layers are less affected by an increased perturbation budget, with BERT less prone than RoBERTa to maintaining the internal consistency of its representation of the original and perturbed sentences (Figure~\ref{fig:similarity-perfdrop}, top-left). In word embeddings (Figure~\ref{fig:similarity-perfdrop}, bottom-left), while the trends of cosine similarity between GloVe and Word2Vec are similar, the $\ell_2$-norm of Word2Vec does not change as much as for GloVe, a sign that in this representation words lie very close to each other w.r.t. the Euclidean distance.
When it comes to performance drop (Figure~\ref{fig:similarity-perfdrop}, right), static and dynamic representations are comparable as they are both not worse than
those
induced by a single coPOS perturbation, i.e., with $\tau=1$.
The performance drop on the coCO and the baseline method are reported in the Appendix, Figures~\ref{fig:similarity-perfdrop-coco} and~\ref{fig:similarity-perfdrop-baseline}.

\begin{figure*}
     \centering
     \includegraphics[width=0.80\linewidth]{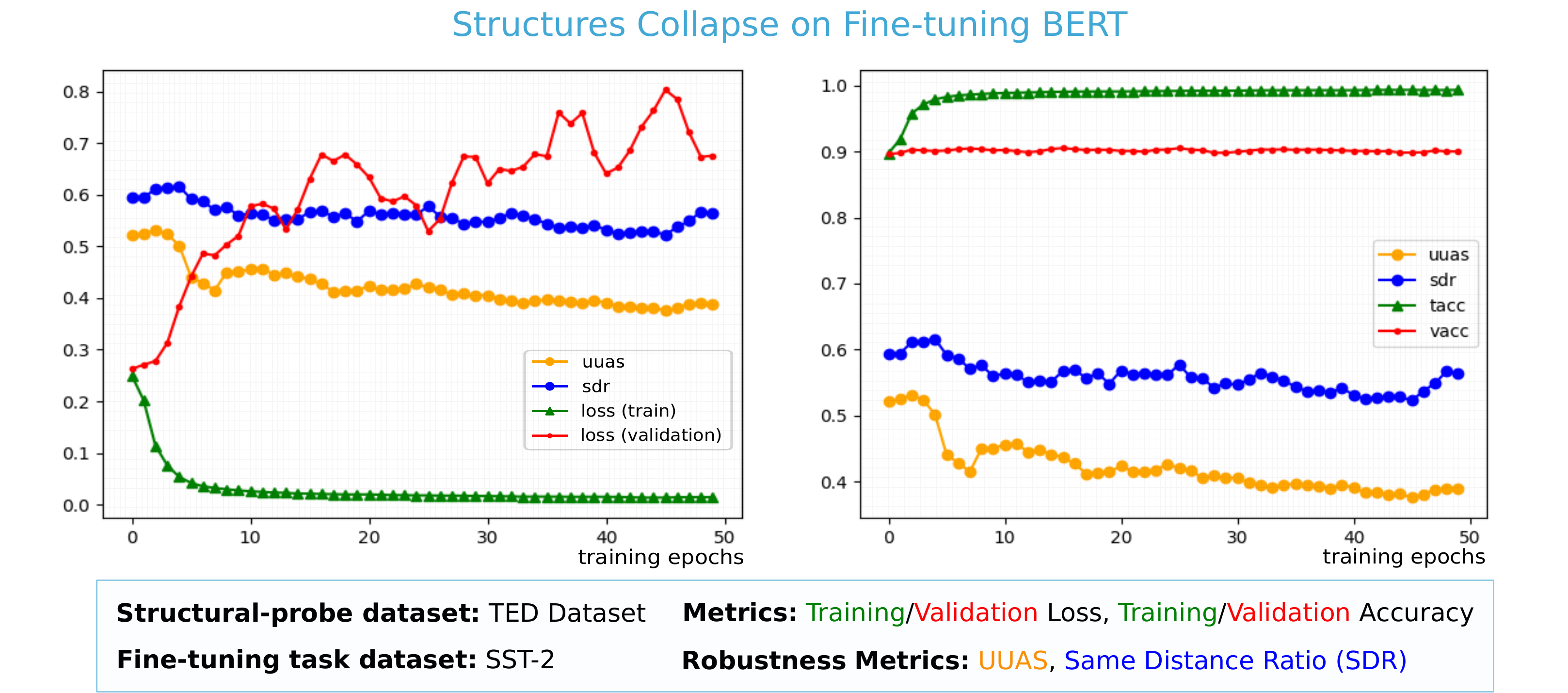}
     \caption{BERT model fine-tuned (and finally, overfitted) on the SST dataset, while its representations are used to train a model to solve the structural probe task. Train and validation losses (left) and accuracies (right) pertain to the fine-tuning task (SST), while %
     SDR and UUAS show the performance on the structural probe. The syntactic metrics degrade as the fine-tuning process proceeds, yet severe over-fitting does not harm syntactic structures.}
     \label{fig:syntax-collapse}
\end{figure*}

\paragraph{On linear vs. non-linear probes}
In our experiments, we use ReLU-activated probing tasks, which constitute a stronger model than classical linear probes~\cite{conneau2018you}. There have been arguments recently %
in favour of non-linear probes, motivated by the fact that linguistic structures are not necessarily encoded linearly by linguistic representations~\cite{pimentel2020information,white2021non}. While we choose to report the results that concern non-linear probes, we conducted the same experiments with linear probes, for which results are released in the code.
\par 
Experiments with linear probes are meant to (i) provide a robustness comparison to the ReLU setting; and (ii) contribute to the debate on the effectiveness of linear probes as tools to investigate the internals of LLMs, since non-linear probes attached to linguistic representations may are on one hand more accurate, yet can incur in overfitting and potentially hinder the expressiveness capabilities of a linguistic representation~\cite{niven2019probing}. 

\paragraph{The effect of fine-tuning and overfitting on syntactic structures}\looseness=-1 
We finally conduct an analysis whose primary intent is to understand the effect of counter-fitting \cite{mrkvsic2016counter} and fine-tuning on syntactic structures, respectively for context-free and context-dependent representations. We train a counter-fitted version of GloVe and we fine-tune, and finally overfit, a BERT-based representation on the SST-2 dataset~\cite{socher2013recursive}: differently from previous experiments, we update the weights of the language model (i.e., we do not keep them frozen) to investigate the existence of some form of catastrophic forgetting of the syntactic structures encoded in the model.
By performing the robustness analysis introduced in this paper, we observe that fine-tuning negatively affects the performance of both shallow and deep context-dependent representations; despite this, excessive fine-tuning is not significantly harmful as the performance does not collapse even after many epochs the model has overfitted on the training set. The task's validation loss is informative to prevent overfitting on the structural probe task, while accuracy on the fine-tuning task can be misleading and hide syntax collapse. We sketch the training dynamics, along with the accuracy of the model on the classification task and the structural probe metrics, in Figure \ref{fig:syntax-collapse}.  
We also observe that any metric of a counter-fitted model has inferior performance on any task and any dataset while being equally brittle to coPOS perturbations, thus limiting the utility of counter-fitting on models aimed at capturing different aspects of human linguistics: performance and robustness are in line with those of standard GloVe embedding (i.e., Table~\ref{tab:distance} and Figure~\ref{fig:per-rob-t12-copos}); we report an extended evaluation of this phenomenon in the Appendix.

\paragraph{Justification for the linguistic structures collapse}
In light of the empirical evidence provided in this paper, we conclude with some hypotheses on why good performance of linguistic representations, whether LLMs or standard word embeddings, comes at the cost of brittleness on high-order syntactic tasks. Certainly, the robustness-performance trade-off accounts for the frailty of over-fitted probes~\cite{madry2017towards}. On the other hand, the absence \textit{stricto sensu} of adversarial attacks, replaced by coPOS and coCO perturbations, forces us to second-guess the existence of such structures. In high-dimensional spaces, vectors (i.e., words) become progressively harder to distinguish, while at the same time the high-dimensionality allows one to optimize a decision boundary that is overfitted on the training set, but fails poorly on slight variations of an input. The hypothesis that we put forward in this paper, with the aim to stimulate discussion among NLP researches as well as linguists, is that linguistic structures emerge as a process of fitting between static sentences and their syntax trees, granted by rich linguistic representations which nonetheless collapse as soon as the input distribution allows for word substitution, a shift against which human linguistic structures are indeed extremely robust.

\section{Conclusion and Future Works}
In this work we studied a notion of syntactic robustness for linguistic representations. Robustness is a desirable property of such models, yet we have exhibited the risk of taking  %
their inherent robustness for granted. We gave empirical evidence of severe brittleness of both LLMs and word embeddings when perturbed via the coPOS and coCO method with restrained perturbation budget. We complemented this by an analysis of the dynamics of robustness w.r.t. overfitting and counter-fitting.
In future, we aim to extend this empirical framework to tasks where the syntactic information is not explicitly provided, and further exploring formally such `syntactic subspaces' where common linguistic features lie: the aim is the foundation of a `linguistics for Large Language Models'.

\section{Acknowledgements}
This project has received funding from the European Research Council (ERC)
under the European Union’s Horizon 2020 research and innovation programme
(FUN2MODEL, grant agreement No.~834115).

\bibliography{anthology,custom}
\bibliographystyle{plain}

\newpage
\appendix

\section{Algorithm for Evaluating Robustness}\label{app:algorithm}
\paragraph{Average distance of the farthest representation} As a measure of distance between sentences, we rely on the $\ell_2$-norm and the cosine similarity, whose usage is widespread in NLP robustness~\cite{huang2019achieving,jia2019certified}, noting that other measure are also possible~\cite{dong2021towards,kusner2015word,la2021guaranteed}. We provide a sketch of the procedure in Alg.~\ref{alg:eps-robustness}. It estimates the average distance of the farthest perturbation in the representation space of $\psi^{\theta}$, w.r.t. a measure of distance like an $\ell$-p norm. The procedure easily accounts for measures of similarity, like the cosine similarity, by changing $max$ with $min$ at line~\ref{alg1:worst}.

\paragraph{Average worst-case syntactic robustness} 
Complementary to Algorithm~\ref{alg:eps-robustness}, Algorithm~\ref{alg:syntax-robustness} permits to evaluate syntax robustness of a linguistic representation $\psi^{\theta}$.
For each pair of a model and a probing task ($f_i$, $\mathbf{T_i}$), we draw a pool of sentences $S_i$ from a CONLL corpus of choice (lines \ref{alg:ptasks}, \ref{alg:getdata}); then, for each sentence $s \in S_i$, we compute a set of coPOS perturbations (lines \ref{alg:budget-k}, \ref{alg:getdatasub}). 
The ratio between the number of sentences in $S_i$ and the perturbations depends on the budget parameter $k$, as well as the number of words per sentence $\tau$ that are perturbed; e.g., with $\tau=k=1$, each sentence in $S_i$ is perturbed once via a single-word substitution. We rely on WordNet \cite{miller1998wordnet} and its graph of synonyms to draw, for each sentence $s$, a substitute $s'$ that is syntax-preserving. We exemplify this process is Figure \ref{fig:running-example}. 
We then quantify the drop of performances of $f_i$ on the original vs. perturbed input representations via the performance measure $\mathcal{L}_i$ (line \ref{alg:perf}).
As we aim for a measure of robustness against perturbations, we return for each sentence $s \in S_i$ the worst-case drop induced by any of the $s'$ generated previously, averaged over the number of test cases (lines \ref{alg:max}, \ref{alg:max1} and \ref{alg:avg}).
We can now pair the measure of robustness with the $\epsilon$-distance between $S$ and the set of worst-case perturbations $S'$ (Def. \ref{def:eps-dist}) as the largest deviation of a pair of input/perturbation w.r.t. the representation $\psi^{\theta}$.

\begin{algorithm} 
\caption{Estimate the average distance of the farthest perturbations w.r.t. a representation $\psi^{\theta}$.}\label{alg:eps-robustness}
\begin{algorithmic}[1]
\Require $\psi^{\theta}(\cdot), S, sub(\cdot), k, dist(\cdot, \cdot)$
\Ensure Average distance of the farthest representation of $\psi^{\theta}_{s\sim S}(s)$ against $sub_{s' \sim S'}(s')$
\State $rob = 0.$
\For{$s \in S$}
    \State $x \gets \psi^{\theta}(s)$
    \State $worst = 0$
        \For{$j$ \textbf{in} $[1,..,k]$}
            \State $s' \gets sub(s)$
            \State $x' \gets \psi^{\theta}(s')$ \Comment{Obtain the representation of a perturbed input}
            \State $d = dist(x, x')$ \Comment{Calculate the distance between input and perturbation}
            \State $worst=max(worst, d)$ \Comment{Worst-case as farthest perturbation}\label{alg1:worst}
        \EndFor
    \State $rob \mathrel{+}= worst$ 
\EndFor
\State \textbf{return} $rob/|S|$ \Comment{Average over each worst-case.}
\end{algorithmic}
\end{algorithm}

\begin{algorithm} 
\caption{Estimate the average worst-drop of robustness of $\psi^{\theta}$ on probing tasks $\mathbf{T}$.}\label{alg:syntax-robustness}
\begin{algorithmic}[1]
\Require $\psi^{\theta}(\cdot), \{\mathbf{T_1}, .., \mathbf{T_m}\}, \{f_1(s), .., f_m(s)\},  \{\mathcal{L}_1, .., \mathcal{L}_m\},  sub(\cdot), \tau, k$
\Ensure Average worst-drop of robustness of $\psi^{\theta}_{s\sim S}(s)$ against $sub_{s' \sim S'}(s')$ on each task $\{\mathbf{T_1}, .., \mathbf{T_m}\}$
\State $Drop = \{\}$ \Comment{Will contain the average worst-case drop per task $\mathbf{T}_i$}
\For{$i \in [1,..,m]$} \label{alg:ptasks}
    \State $drop = 0.$ \Comment{Average worst-case drop of robustness}
    \State $S_i \gets data(\mathbf{T}_i)$ \label{alg:getdata} \Comment{Get data from each task}
    \For {$s \in S_i$}
        \State $d = 0.$
        \For{$j$ \textbf{in} $[1,..,k]$} \label{alg:budget-k}
            \State $s' \gets sub(s, \tau)$ \label{alg:getdatasub} \Comment{$\tau$ words are perturbed to obtain $s'$ from $s$}
            \State $x, x' \gets \psi^{\theta}(s), \psi^{\theta}(s')$ \Comment{Input/perturbation pairs}
            \State $\Delta d = \mathcal{L}_i(f_i(x), f_i(x'))$ \label{alg:perf} \Comment{Drop of robustness between input and perturbation}
            \State{$d = max(d, \Delta d)$} \label{alg:max} \Comment{Get the case that minimizes syntax robustness}
        \EndFor
        \State $drop \mathrel{+}= d$ \label{alg:max1}
    \EndFor
    \State $Drop \xleftarrow{+} drop/|S_i|$ \label{alg:avg}
\EndFor
\State \textbf{return} $Drop$
\end{algorithmic}
\end{algorithm}

\section{Experiments}
As described in the experimental section, we perform our analyses on 4 linguistic representations, namely GloVe~\cite{pennington2014glove}, Word2Vec~\cite{mikolov2013distributed}, BERT~\cite{devlin2018bert} and RoBERTa~\cite{liu2019roberta}.

We searched for the best architecture in terms of performance on the four tasks on the 6 corpora of interest: we found that the convolutional architecture performs poorly on the tasks, while  recurrent architectures are competitive with fully connected, but introduce additional computation overhead and the intrinsic inductive bias of the recurrent gates.

We applied the least amount of pre-processing to the input texts, i.e., we split sentences into words -- both for the context-free and the context-dependent representations. As the size of the input matrix for the \textit{syntax reconstruction} task grows quadratically with the input length, we cut (pad) the sentences longer (shorter) than 20 words. 

As regards the training, models have been trained until they start overfitting the data, i.e., we end the procedure via early-stop.

For further details on the architectures, e.g., the number of parameters, the initialization, etc., we provide the logs of each experiment as part of the code to reproduce the experiments.

\section{Logs for all the experiments of the paper (and more)}
We make the code available, which alongside the instructions to run it can be found at \url{https://github.com/EmanueleLM/emergent-linguistic-structures}. We also provide all the logs to reconstruct the results we present in this paper. Files are stored as plain text in the \texttt{.zip} archive, under the \\
\noindent\texttt{{robust-linguistic-structures}\textbackslash verify\textbackslash MLM\_internals\textbackslash syntax-integrity \textbackslash results} folder.

\section{Experiments: Additional Results}

\subsection{On Spearman correlation's drop as a measure of robustness}\label{app:spearman}
\begin{figure*}
     \centering
     \includegraphics[width=1.\linewidth]{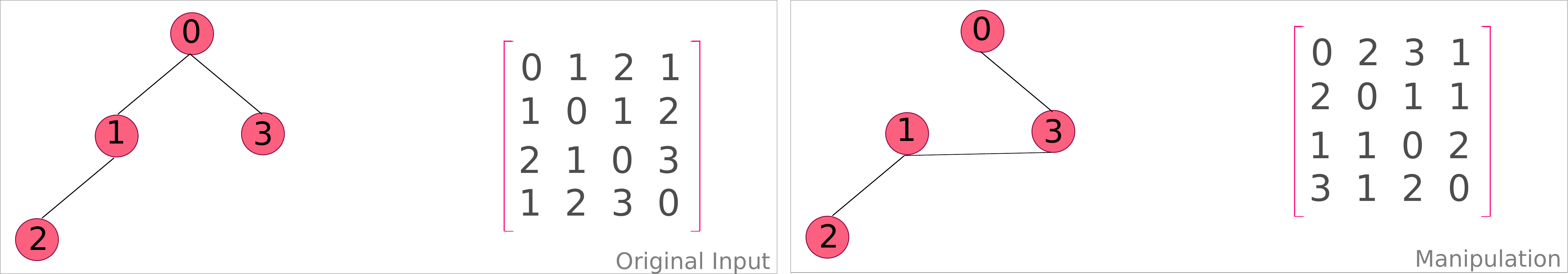}
     \caption{Example of Spearman correlation and its lack of consistency when judging syntax manipulations.}
     \label{fig:spearman-example}
\end{figure*}

While the Spearman correlation metric is used in~\cite{manning2020emergent} to measure the capacity of a model to represent a sentence's syntax tree, we found two reasons why that this metric is not indicative of the robustness of a model. To illustrate the first reason, we present an example: consider the syntax tree presented in Figure~\ref{fig:spearman-example} (left), which encodes the distances between the sentence $[0,1,2,3]$. Its distance metric between nodes is also reported in Figure~\ref{fig:spearman-example} (right). Now, suppose that a perturbation makes the tree change from that %
on the left to that on the right, with the relative distance matrix reported on the right. The Spearman coefficient between the two trees is, according to a standard implementation\footnote{\url{https://docs.scipy.org/doc/scipy/reference/generated/scipy.stats.spearmanr.html}}, $0.65$; however, %
we could not consider a model, which outputs the second tree as a result of a sentence manipulation, robust as the difference between the syntax structures is not reflected in the value of the coefficients.
\newline 
Furthermore, we discovered that for all the experiments, both the coPOS and coCo perturbation methods induce a shift of each word-to-word distance towards negative values. One might argue that the shift can preserve the information of the original tree, e.g., any word-to-word distance is for example shifted by a negative constant $c$: in that case, one could revert the predicted tree with a simple mathematical operation. Unfortunately, this is not possible, as the shift varies sensibly from word to word while maintaining the correlation between the original and the predicted trees high.

\subsection{coCO and Baseline Perturbations}
\begin{figure*}
     \centering
     \includegraphics[width=1.\linewidth]{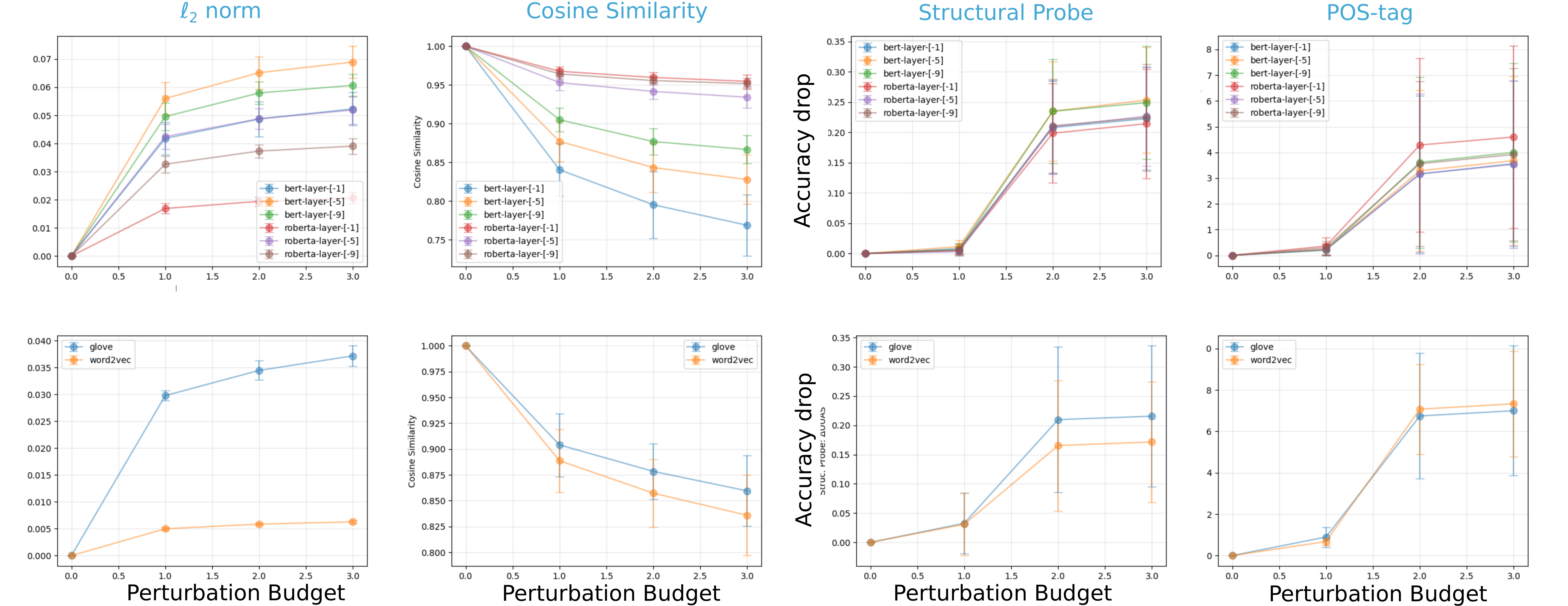}
     \caption{
     Left: for an increasing perturbation budget $\tau$ and the coco method, cosine similarity between perturbed and original sentences drops, while the $\ell_2$-norm increases. Right: It is clear that, even with $\tau=2$ (i.e., at most two words per-sentence are perturbed), the models' performance already experiences a significant drop (the higher the curve, the worse the model is on a syntactic task). Increasing the perturbation budget does not lead to a significant drop of robustness. 
     }
     \label{fig:similarity-perfdrop-coco}
\end{figure*}

\begin{figure*}
     \centering
     \includegraphics[width=1.\linewidth]{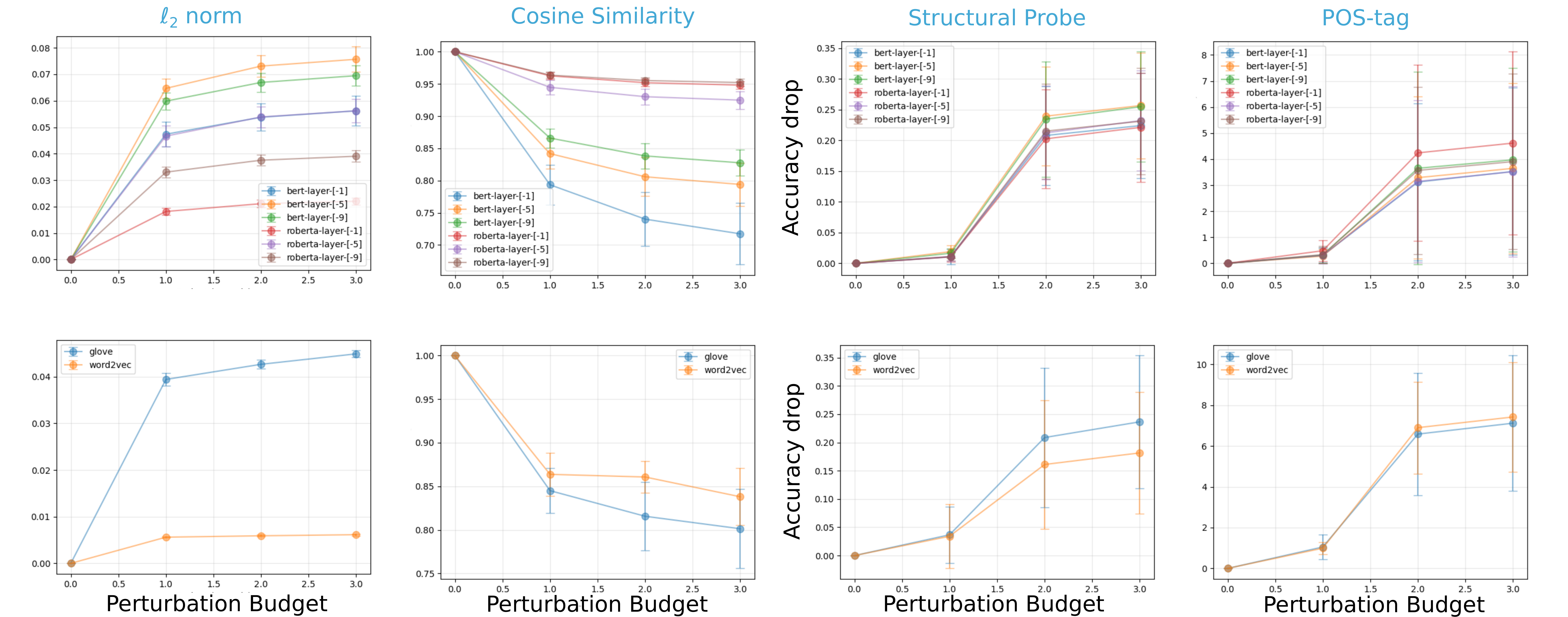}
     \caption{
     Left: For an increasing perturbation budget $\tau$ and the baseline method, cosine similarity between perturbed and original sentences drops, while the $\ell_2$-norm increases. Right: It is clear that, even with $\tau=2$ (i.e., at most two words per-sentence are perturbed), the models' performance already experiences a significant drop (the higher the curve, the worse the model is on a syntactic task). Increasing the perturbation budget does not lead to a significant drop of robustness. 
     }
     \label{fig:similarity-perfdrop-baseline}
\end{figure*}

In Figure~\ref{fig:similarity-perfdrop-coco}, we report the cosine similarity and the $\ell_2$ drop between pairs of inputs and perturbations (left), and the relative drop of performances on the probing tasks (right), when inputs are targeted by coCO perturbations. The same results, but relative to the baseline perturbation method, are reported in Figure~\ref{fig:similarity-perfdrop-baseline}. Results are comparable to those with coPOS perturbations, thus strengthening the hypothesis that linguistics structures, if present, are brittle to slight, syntax-preserving perturbations.

\subsection{The Effect of Fine-tuning and Over-fitting on Syntactic Structures}
We report the performances of GloVe counter-fitted models on the four probing tasks in Tables \ref{tab:glove-counterfitted-bert-finetuned} and~\ref{tab:glove-counterfitted-bert-finetuned-coco}.
As it can be observed, the performance on each task is inferior to standard GloVe embedding, suggesting that counter-fitting is harmful to the syntactic structures encoded in the representation.
When targeted with perturbations, the syntactic structures of GloVe counter-fitted models collapse,  thus, we can conclude that counter-fitting does not improve the syntactic capabilities of a model, yet it is equivalently as brittle as GloVe. 

\begin{table*}[]\footnotesize
\centering
{{\textbf{\ref{tab:glove-counterfitted-bert-finetuned-coco}.1 Robustness}}}
\vspace{1 mm}
\\
\begin{tabular}{|l|c|c|c|c|}
\hline
    \multirow{2}{2cm}{\textbf{}} & \multicolumn{3}{c|}{\textbf{Syntax Reconstruction}} & \textbf{POS-tagging}\\
    \cline{2-5}
    & \textbf{$\Delta$SDR} & \textbf{$\Delta$UUAS} & \textbf{$\Delta$Sp} & \textbf{$\Delta$Acc.}\\ \hline
\textbf{TED}             & $0.167$ & $0.0872$ & $0.0011$ &             $4.470$     \\ \hline
\textbf{En-Universal}          & $0.2360$ & $0.29931$ & $0.0056$ &             $9.456$     \\ \hline
\textbf{Ud-English-ewt}    & $0.1463$ & $0.3124$ & $0.0.106$ &                 $6.951$     \\ \hline
\textbf{Ud-English-gum} & $0.1678$ & $0.3252$ & $0.0025$ &                 $3.253$     \\ \hline
\textbf{Ud-English-lines}    & $0.2599$ & $0.2981$ & $0.0022$ &                 $7.402$     \\ \hline
\textbf{Ud-English-pud} & $0.0574$ & $0.0.0612$ & $0.0010$ &                 $5.746$     \\ \hline
\end{tabular}
\vspace{2 mm}
\\
{{\textbf{\ref{tab:glove-counterfitted-bert-finetuned-coco}.2 Robustness}}}
\\
\vspace{1 mm}
\begin{tabular}{|l|c|c|c|c|}
\hline
    \multirow{2}{2cm}{\textbf{}} &  \textbf{Root Identification} & \multicolumn{2}{c|}{\textbf{Tree Depth Estimation}}\\
    \cline{2-4}
    & \textbf{$\Delta$Acc.} & \textbf{$\Delta$Acc.} & \textbf{$\Delta$Sp}\\ \hline
\textbf{TED}             & $0.3753$ &             $0.0727$ & $0.0165$     \\ \hline
\textbf{En-Universal}          & $0.5288$ &             $0.2215$ & $-0.0011$     \\ \hline
\textbf{Ud-English-ewt}    & $0.816$ &                 $0.7751$ & $0.0102$     \\ \hline
\textbf{Ud-English-gum} & $0.389$ &                 $0.3065$ & $0.0028$     \\ \hline
\textbf{Ud-English-lines}    & $0.4493$ &                 $0.314$ & $0.0037$     \\ \hline
\textbf{Ud-English-pud} & $0.3894$ &                 $0.2583$ & $-0.0189$     \\ \hline
\end{tabular}
\vspace{2 mm}
\\
{{\textbf{\ref{tab:glove-counterfitted-bert-finetuned-coco}.3 Distance/Similarity Metrics}}}
\\
\vspace{1 mm}
\begin{tabular}{|l|c|c|}
\hline
                  & \textbf{Cosine similarity}  & \textbf{$\ell_2$-norm distance} \\ \hline
\textbf{TED}             & $0.881$ & $0.006$ \\ \hline
\textbf{En-Universal}          & $\approx 1.$ & $0.0033$   \\ \hline
\textbf{Ud-English-ewt}    & $0.9295$ & $0.005$      \\ \hline
\textbf{Ud-English-gum} & $\approx 1.$ & $0.005$     \\ \hline
\textbf{Ud-English-lines}    & $\approx 1.$ & $0.0058$     \\ \hline
\textbf{Ud-English-pud} & $0.881$ & $0.0058$     \\ \hline
\end{tabular}
\caption{
Robustness of GloVe counter-fitted models, with an analysis, w.r.t. each dataset (one per row), of the relationship between the syntactic robustness metrics for coCO perturbations with budget $\tau=2$ (top and middle) and the distance between pairs of perturbations and original inputs (bottom). %
The accuracy drop of the POS-tag task is reported in number of words correctly guessed. Results confirm that counter-fitting does not improve robustness at any level (in this case, syntactic robustness).
}
\label{tab:glove-counterfitted-bert-finetuned-coco}
\end{table*}

\begin{table*}[]\footnotesize
\centering
{{\textbf{\ref{tab:glove-counterfitted-bert-finetuned}.1 Robustness}}}
\vspace{1 mm}
\\
\begin{tabular}{|l|c|c|c|c|}
\hline
    \multirow{2}{2cm}{\textbf{}} & \multicolumn{3}{c|}{\textbf{Syntax Reconstruction}} & \textbf{POS-tagging}\\
    \cline{2-5}
    & \textbf{$\Delta$SDR} & \textbf{$\Delta$UUAS} & \textbf{$\Delta$Sp} & \textbf{$\Delta$Acc.}\\ \hline
\textbf{TED}             & $0.167$ & $0.0872$ & $0.0011$ &             $4.470$     \\ \hline
\textbf{En-Universal}          & $0.2360$ & $0.29931$ & $0.0056$ &             $9.456$     \\ \hline
\textbf{Ud-English-ewt}    & $0.1463$ & $0.3124$ & $0.0.106$ &                 $6.951$     \\ \hline
\textbf{Ud-English-gum} & $0.1678$ & $0.3252$ & $0.0025$ &                 $3.253$     \\ \hline
\textbf{Ud-English-lines}    & $0.2599$ & $0.2981$ & $0.0022$ &                 $7.402$     \\ \hline
\textbf{Ud-English-pud} & $0.0574$ & $0.0.0612$ & $0.0010$ &                 $5.746$     \\ \hline
\end{tabular}
\vspace{2 mm}
\\
{{\textbf{\ref{tab:glove-counterfitted-bert-finetuned}.2 Robustness}}}
\\
\vspace{1 mm}
\begin{tabular}{|l|c|c|c|c|}
\hline
    \multirow{2}{2cm}{\textbf{}} &  \textbf{Root Identification} & \multicolumn{2}{c|}{\textbf{Tree Depth Estimation}}\\
    \cline{2-4}
    & \textbf{$\Delta$Acc.} & \textbf{$\Delta$Acc.} & \textbf{$\Delta$Sp}\\ \hline
\textbf{TED}             & $0.3753$ &             $0.0727$ & $0.0165$     \\ \hline
\textbf{En-Universal}          & $0.5288$ &             $0.2215$ & $-0.0011$     \\ \hline
\textbf{Ud-English-ewt}    & $0.816$ &                 $0.7751$ & $0.0102$     \\ \hline
\textbf{Ud-English-gum} & $0.389$ &                 $0.3065$ & $0.0028$     \\ \hline
\textbf{Ud-English-lines}    & $0.4493$ &                 $0.314$ & $0.0037$     \\ \hline
\textbf{Ud-English-pud} & $0.3894$ &                 $0.2583$ & $-0.0189$     \\ \hline
\end{tabular}
\vspace{2 mm}
\\
{{\textbf{\ref{tab:glove-counterfitted-bert-finetuned}.3 Distance/Similarity Metrics}}}
\\
\vspace{1 mm}
\begin{tabular}{|l|c|c|}
\hline
                  & \textbf{Cosine similarity}  & \textbf{$\ell_2$-norm distance} \\ \hline
\textbf{TED}             & $0.925$ & $0.0037$ \\ \hline
\textbf{En-Universal}          & $0.919$ & $0.0033$   \\ \hline
\textbf{Ud-English-ewt}    & $0.9295$ & $0.0035$      \\ \hline
\textbf{Ud-English-gum} & $\approx 1.$ & $0.0036$     \\ \hline
\textbf{Ud-English-lines}    & $\approx 1.$ & $0.0037$     \\ \hline
\textbf{Ud-English-pud} & $0.9518$ & $0.0038$     \\ \hline
\end{tabular}
\caption{
Robustness of GloVe counter-fitted models, with an analysis, w.r.t. each dataset (one per row), of the relationship between the syntactic robustness metrics for coPOS perturbations with budget $\tau=2$ (top and middle) and the distance between pairs of perturbations and original inputs (bottom). %
The accuracy drop of the POS-tag task is reported in number of words correctly guessed. Results confirm that counter-fitting does not improve robustness at any level (in this case, syntactic robustness).
}
\label{tab:glove-counterfitted-bert-finetuned}
\end{table*}

\begin{figure*}
     \centering
     \includegraphics[width=1\linewidth]{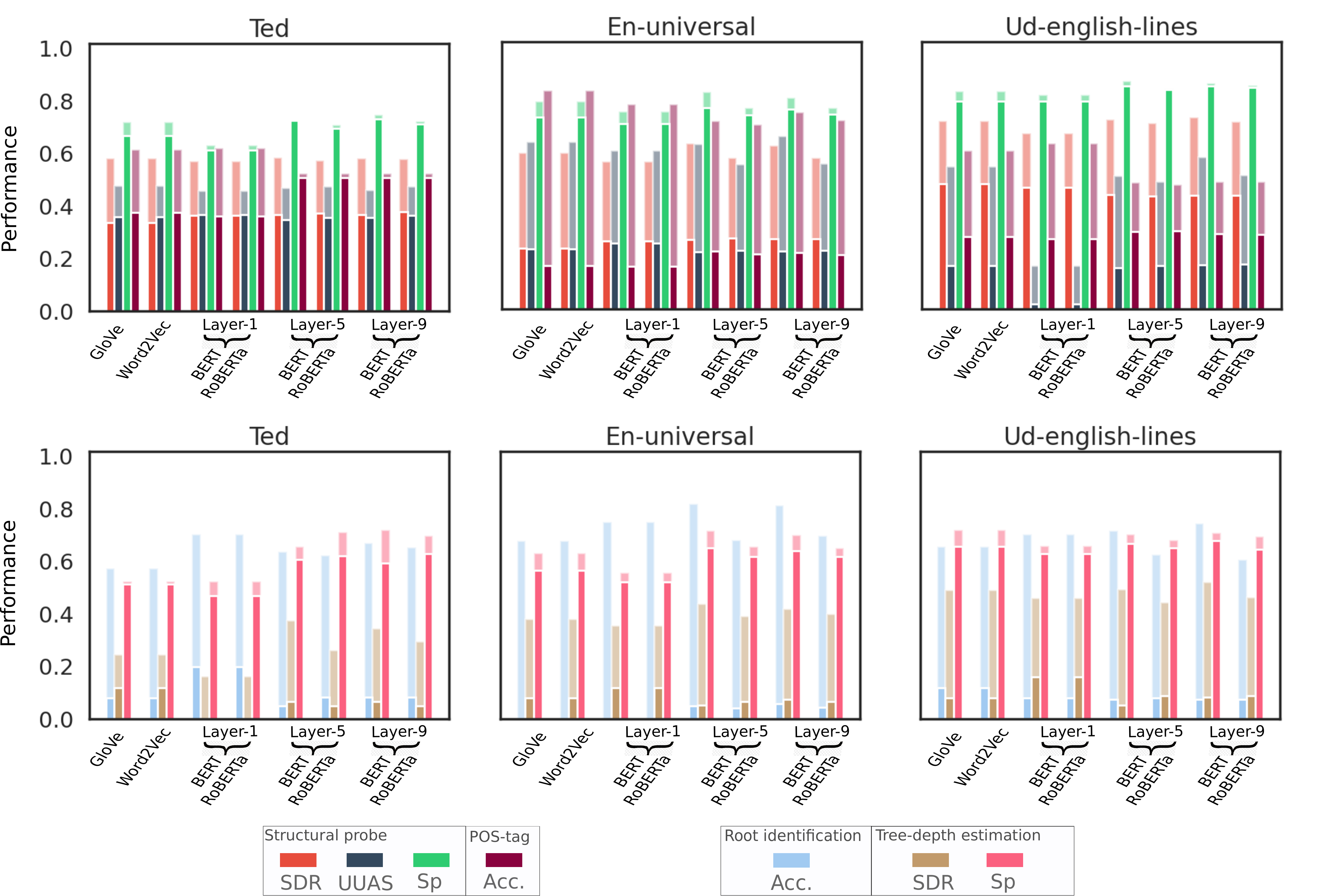}
     \caption{Performance, against baseline perturbations, of different linguistic representations on \textit{syntax reconstruction} and \textit{POS-tagging} probing tasks (top) and on \textit{root identification} (accuracy metric) and \textit{tree-depth estimation} (SDR and Spearman metric) probing tasks (bottom). For all plots, the performance of the probing tasks is reported as shaded bars, with the performance for the perturbed representation shown %
     as a solid overlapping bar: the results refer to the case where the baseline perturbation budget $\tau$ is equal to 3 (i.e., at most 3 words per-sentence are perturbed). 
     Regardless of the embedding/representation employed, we observe severe brittleness of the syntactic representations. %
     }
     \label{fig:per-rob-t12-baseline}
\end{figure*}

\begin{figure*}
     \centering
     \includegraphics[width=1\linewidth]{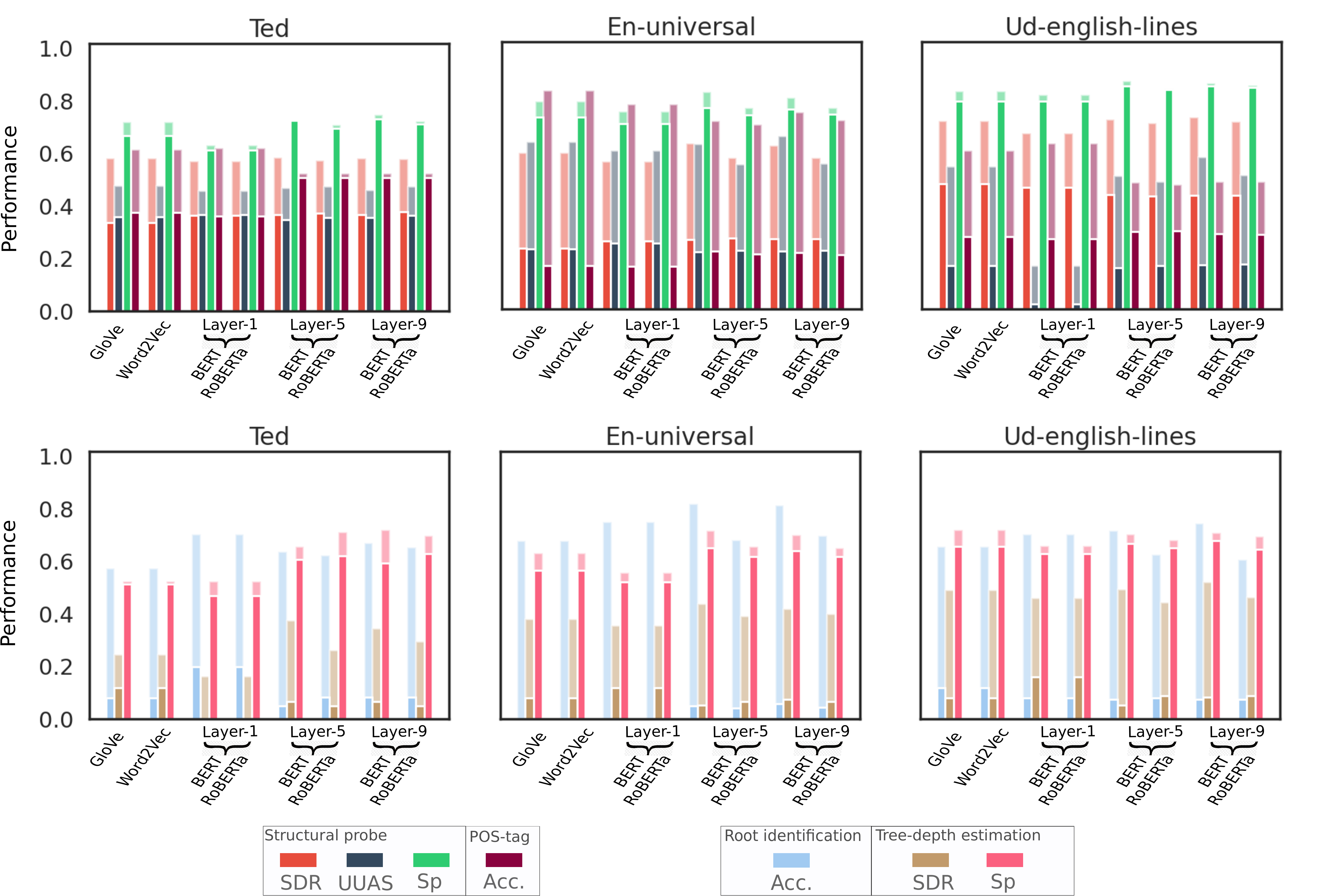}
     \caption{Performance, against coCO perturbations, of different linguistic representations on \textit{syntax reconstruction} and \textit{POS-tagging} probing tasks (top) and on \textit{root identification} (accuracy metric) and \textit{tree-depth estimation} (SDR and Spearman metric) probing tasks (bottom). For all plots, the performance of the probing tasks is reported as shaded bars, with the performance for the perturbed representation shown %
     as a solid overlapping bar: the results refer to the case where the coCO perturbation budget $\tau$ is equal to 3 (i.e., at most 3 words per-sentence are perturbed). 
     Regardless of the embedding/representation employed, we observe severe brittleness of the syntactic representations. %
     }
     \label{fig:per-rob-t12-coco}
\end{figure*}

\begin{figure*}
     \centering
     \includegraphics[width=1\linewidth]{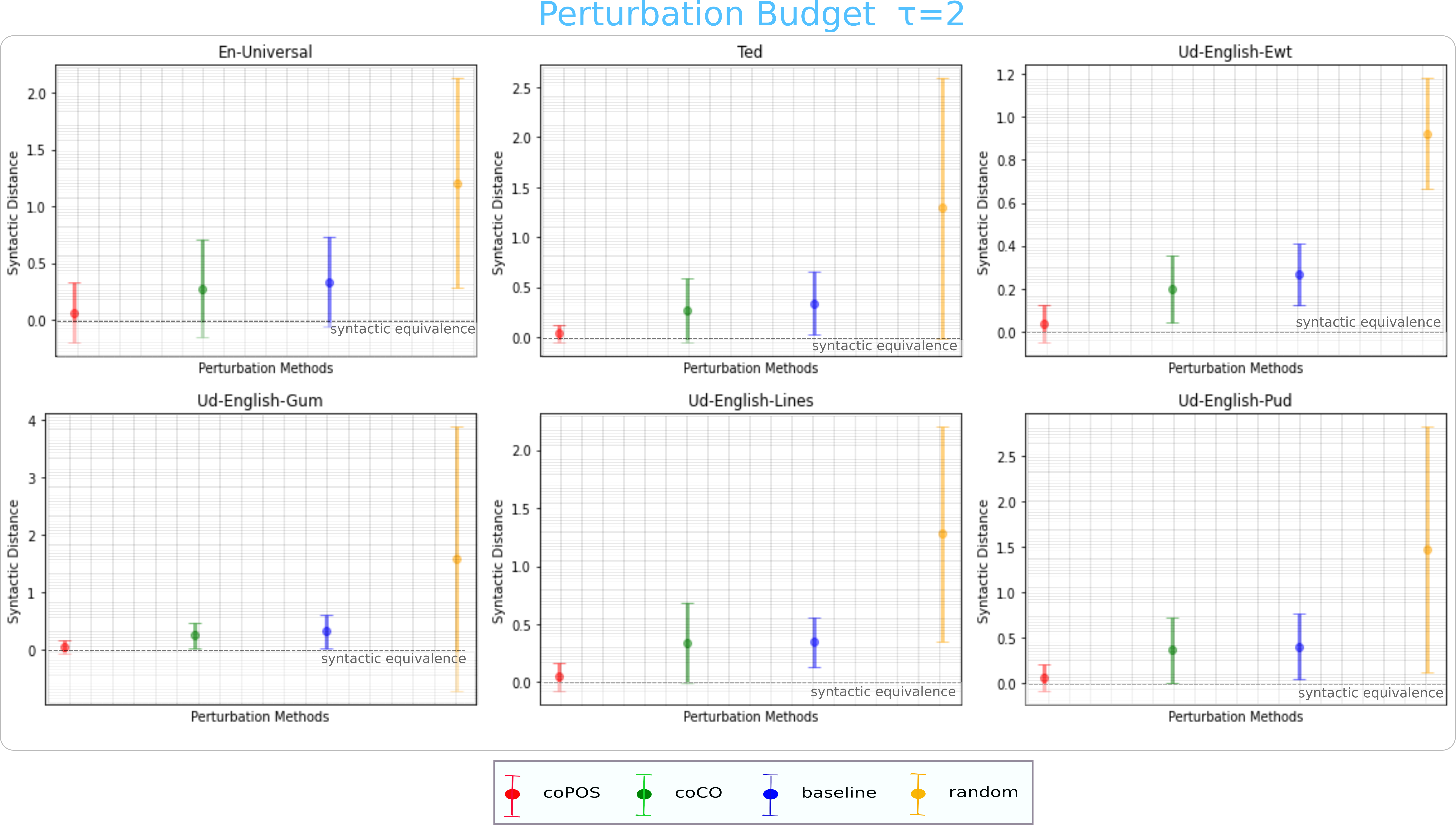}
     \caption{
     Tree-distance, measured with Stanza, between an input and its perturbed version, for different datasets and perturbation budget $\tau=2$. The coPOS perturbation method (\textcolor{red}{red}) induces almost no disruption to a perturbation's syntax tree, while injection of random words (\textcolor{blue}{blue}) and coPOS perturbations (\textcolor{greentxt}{green}) both induce some noticeable disruption. The disruption induces by comparing the syntax tree of two random sentences is reported for comparison (\textcolor{orange}{orange}).
     }
     \label{fig:validate-mean-std-tau2}
\end{figure*}

\begin{table*}[]\footnotesize
\centering
{{\textbf{\ref{tab:distance-tau2}.1 Robustness}}}
\vspace{1 mm}
\\
\begin{tabular}{|l|c|c|c|c|}
\hline
    \multirow{2}{2cm}{\textbf{}} & \multicolumn{3}{c|}{\textbf{Syntax Reconstruction}} & \textbf{POS-tagging}\\
    \cline{2-5}
    & \textbf{$\Delta$SDR} & \textbf{$\Delta$UUAS} & \textbf{$\Delta$Sp} & \textbf{$\Delta$Acc.}\\ \hline
\textbf{GloVe}            & $0.2087 \pm 0.123$ & $0.2822 \pm 0.1498$ & $0.0545 \pm 0.0097$ &             $6.583 \pm 2.9971$      \\ \hline
\textbf{Word2Vec}          & $0.161 \pm 0.1137$ & $0.139 \pm 0.0996$ & $0.035 \pm 0.0079$ &             $6.8977 \pm 2.2552$     \\ \hline
\textbf{BERT layer -1}     & $0.208 \pm 0.0803$ & $0.1787 \pm 0.0827$ & $0.0289 \pm 0.0156$ &                 $3.1208 \pm 3.0138$     \\ \hline
\textbf{RoBERTa layer -1}  & $0.2026 \pm 0.0805$ & $0.1954 \pm 0.1129$ & $0.013 \pm 0.0154$ &                 $4.2408 \pm 3.3794$     \\ \hline
\textbf{BERT layer -5}     & $0.2395 \pm 0.0804$ & $0.2599 \pm 0.1287$ & $0.023 \pm 0.0196$ &                 $3.289 \pm 3.1118$     \\ \hline
\textbf{RoBERTa layer -5}  & $0.2123 \pm 0.0764$ & $0.2303 \pm 0.1094$ & $0.0137 \pm 0.0098$ &                 $3.1448 \pm 3.1161$     \\ \hline
\textbf{BERT layer -9}     & $0.2345 \pm 0.0938$ & $0.2828 \pm 0.162$ & $0.0204 \pm 0.0126$ &                 $3.6504 \pm 3.7005$     \\ \hline
\textbf{RoBERTa layer -9}  & $0.2151 \pm 0.0773$ & $0.2372 \pm 0.1189$ & $0.009 \pm 0.0106$ &                 $3.568 \pm 3.2097$     \\ \hline
\end{tabular}
\vspace{2 mm}
\\
{{\textbf{\ref{tab:distance-tau2}.2 Robustness}}}
\\
\vspace{1 mm}
\begin{tabular}{|l|c|c|c|c|}
\hline
    \multirow{2}{2cm}{\textbf{}} &  \textbf{Root Identification} & \multicolumn{2}{c|}{\textbf{Tree Depth Estimation}}\\
    \cline{2-4}
    & \textbf{$\Delta$Acc.} & \textbf{$\Delta$Acc.} & \textbf{$\Delta$Sp}\\ \hline
\textbf{GloVe}             & $0.4853 \pm 0.1776$ &             $0.2796 \pm 0.2265$ & $-0.0108 \pm 0.1222$     \\ \hline
\textbf{Word2Vec}          & $0.6118 \pm 0.1342$ &             $0.3115 \pm 0.2495$ & $0.0407 \pm 0.0156$     \\ \hline
\textbf{BERT layer -1}    & $0.5466 \pm 0.2041$ &                 $0.3883 \pm 0.2186$ & $0.103 \pm 0.0784$     \\ \hline
\textbf{RoBERTa layer -1} & $0.5557 \pm 0.1783$ &                 $0.3586 \pm 0.2312$ & $0.0818 \pm 0.0306$     \\ \hline
\textbf{BERT layer -5}    & $0.6466 \pm 0.1421$ &                 $0.3896 \pm 0.2246$ & $0.0317 \pm 0.0575$     \\ \hline
\textbf{RoBERTa layer -5} & $0.5464 \pm 0.1778$ &                 $0.3252 \pm 0.2351$ & $0.0225 \pm 0.0704$     \\ \hline
\textbf{BERT layer -9}    & $0.6462 \pm 0.1392$ &                 $0.3927 \pm 0.2461$ & $0.0314 \pm 0.0464$     \\ \hline
\textbf{RoBERTa layer -9} & $0.5563 \pm 0.1834$ &                 $0.3485 \pm 0.2493$ & $0.0344 \pm 0.046$     \\ \hline
\end{tabular}
\vspace{2 mm}
\\
{{\textbf{\ref{tab:distance-tau2}.3 Distance/Similarity Metrics}}}
\\
\vspace{1 mm}
\begin{tabular}{|l|c|c|}
\hline
                  & \textbf{$\ell_2$-norm distance}  & \textbf{Cosine similarity}  \\ \hline
\textbf{GloVe}             & $0.0427 \pm 0.001$ & $0.8156 \pm 0.0391$     \\ \hline
\textbf{Word2Vec}           &  $0.0059 \pm 0.0002$ & $0.8606 \pm 0.0184$       \\ \hline
\textbf{BERT layer -1}    & $0.0538 \pm 0.0052$ & $0.7401 \pm 0.0418$       \\ \hline
\textbf{RoBERTa layer -1} & $0.0211 \pm 0.0014$ & $0.9519 \pm 0.006$       \\ \hline
\textbf{BERT layer -5}    & $0.0731 \pm 0.0041$ & $0.8059 \pm 0.0299$       \\ \hline
\textbf{RoBERTa layer -5} & $0.0539 \pm 0.004$ & $0.9302 \pm 0.0122$       \\ \hline
\textbf{BERT layer -9}    & $0.0669 \pm 0.0035$ & $0.8382 \pm 0.0199$       \\ \hline
\textbf{RoBERTa layer -9} & $0.0375 \pm 0.0021$ & $0.9553 \pm 0.0051$       \\ \hline
\end{tabular}
\caption{%
Relationship between the syntactic robustness metrics for four linear probing tasks on baseline perturbations with budget $\tau=2$ (top and middle row) and the distance between pairs of perturbed and original inputs measured via cosine similarity and $\ell_2$-norm distance (bottom row). %
The accuracy drop of the POS-tag task is reported as the number of words correctly guessed in both cases.
The reported standard deviation is measured by averaging over the 6 training corpora. Whilst the distance (similarity) between inputs and perturbations is low (high), we observe that all embeddings/representations are %
brittle to syntax-preserving perturbations. 
}
\label{tab:distance-tau2}
\end{table*}

\end{document}